\documentclass[lettersize,journal]{IEEEtran}
\usepackage{amsmath,amsfonts}
\usepackage{algorithmic}
\usepackage{algorithm}
\usepackage{array}
\usepackage[caption=false,font=normalsize,labelfont=sf,textfont=sf]{subfig}
\usepackage{textcomp}
\usepackage{stfloats}
\usepackage{url}
\usepackage{verbatim}
\usepackage{graphicx}
\usepackage{cite}
\usepackage{bbding}
\usepackage{multirow}
\usepackage{graphicx}
\usepackage{booktabs}
\usepackage{makecell}
\usepackage{hyperref}
\usepackage[T1]{fontenc}
\begin{document}

\title{An Efficient Wide-Range Pseudo-3D \\Vehicle Detection Using A Single Camera}

\author{Zhupeng Ye, Yinqi Li, Zejian Yuan
\thanks{Zhupeng Ye, Yinqi Li, and Zejian Yuan are with the Institute of Artificial Intelligence and Robotics, Xi’an Jiaotong University, Xi’an 710049, China (e-mail:yezhupeng@stu.xjtu.edu.cn; liyinqi@stu.xjtu.edu.cn; yuan.ze.jian@xjtu.edu.cn)}}

\markboth{Journal of \LaTeX\ Class Files,~Vol.~14, No.~8, August~2023}%
{Shell \MakeLowercase{\textit{et al.}}: A Sample Article Using IEEEtran.cls for IEEE Journals}

\IEEEpubid{0000--0000/00\$00.00~\copyright~2021 IEEE}

\maketitle

\begin{abstract}
Wide-range and fine-grained vehicle detection plays a critical role in enabling active safety features in intelligent driving systems. However, existing vehicle detection methods based on rectangular bounding boxes (BBox) often struggle with perceiving wide-range objects, especially small objects at long distances. And BBox expression cannot provide detailed geometric shape and pose information of vehicles. This paper proposes a novel wide-range Pseudo-3D Vehicle Detection method based on images from a single camera and incorporates efficient learning methods. This model takes a spliced image as input, which is obtained by combining two sub-window images from a high-resolution image. This image format maximizes the utilization of limited image resolution to retain essential information about wide-range vehicle objects. To detect pseudo-3D objects, our model adopts specifically designed detection heads. These heads simultaneously output extended BBox and Side Projection Line (SPL) representations, which capture vehicle shapes and poses, enabling high-precision detection. To further enhance the performance of detection, a joint constraint loss combining both the object box and SPL is designed during model training, improving the efficiency, stability, and prediction accuracy of the model. Experimental results on our self-built dataset demonstrate that our model achieves favorable performance in wide-range pseudo-3D vehicle detection across multiple evaluation metrics. Our demo video has been placed at \href{https://www.youtube.com/watch?v=1gk1PmsQ5Q8}{https://www.youtube.com/watch?v=1gk1PmsQ5Q8}.
\end{abstract}

\begin{IEEEkeywords}
DW image, Joint constraint loss, Pseudo-3D vehicle detection.
\end{IEEEkeywords}

\section{Introduction}
\IEEEPARstart{W}{ide-range} vehicle detection is a fundamental task in autonomous driving. It can provide a distant and broad view for vehicles, helping them perceive potential collisions in advance and significantly improving the safety and stability of autonomous driving. 
Existing detecting methods \cite{detr3d, li2022bevformer} based on multi-camera images perform well in detecting surrounding vehicles, and other work\cite{chadwick2019distant} incorporates radar data to boost performance in detecting small objects. While successful in the past, these approaches incur high labeling costs for multi-camera images and sensor data.
On the other hand, current object detection methods\cite{6909475, 7485869, 7780460, liu2016ssd, Law_2018_ECCV, tian2019fcos, zhou2019objects, li2022cross, song2022msfanet} typically output axis-aligned BBox as the representation of objects, which can only provide rough vehicle size and localization information, and cannot describe the shape, appearance, and pose of the vehicle.
\par
\begin{figure}[htbp]
    \centering
    \includegraphics[width=0.90\linewidth]{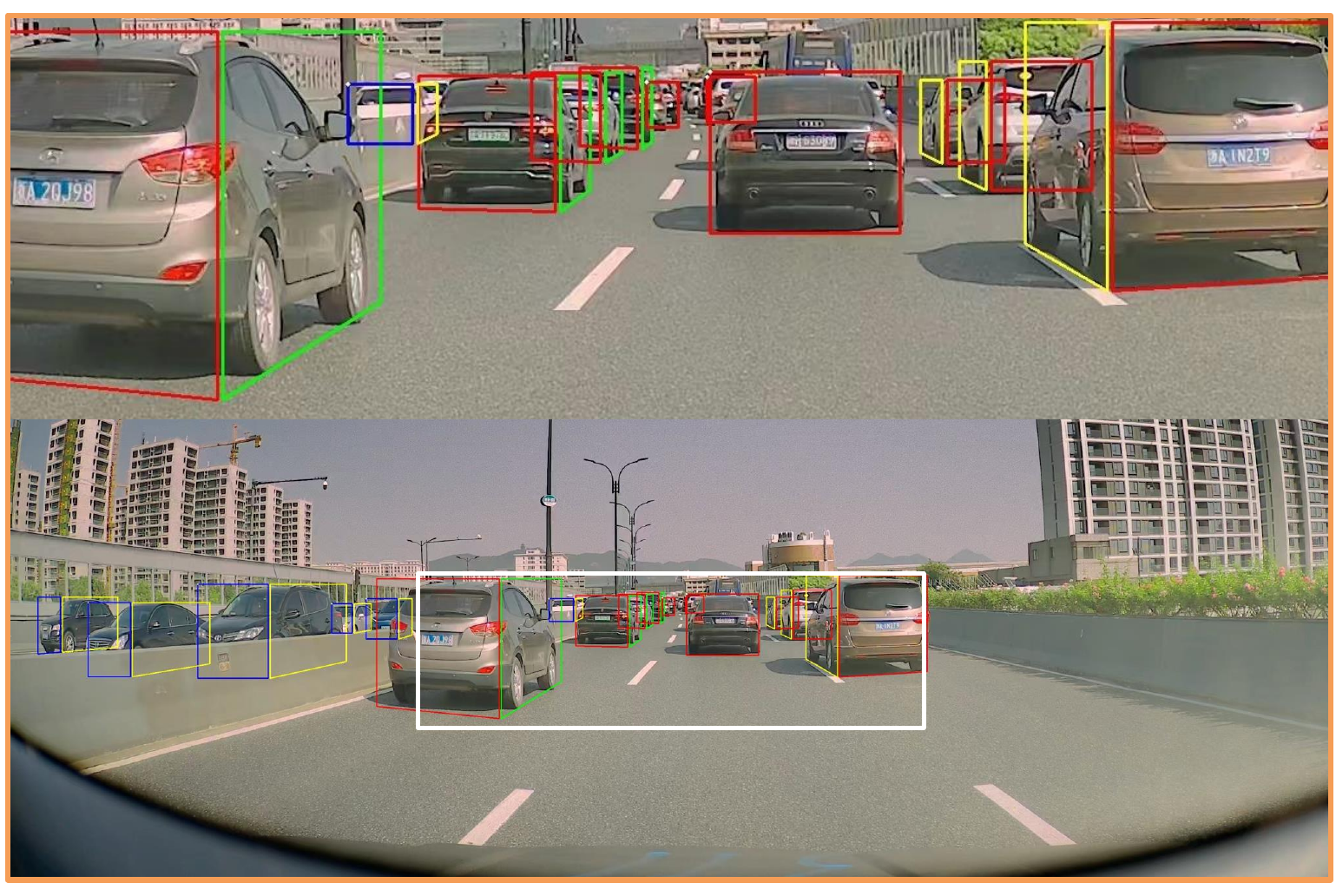}
    \caption{Double Window image and corresponding Pseudo-3D Vehicle Representation detection results. Each DW image is obtained by concatenating Center Window (CW) image and Global Window (GW) image up and down.}
    \label{fig:DW_Visualization}
\end{figure}

This paper proposes an efficient wide-range Pseudo-3D Vehicle Detection (P3DVD) method based on images from a single camera, which combines high-precision detection for small distant targets and high detecting efficiency. Our method uses Double-Window (DW) images as the dataset, as shown in Fig. \ref{fig:DW_Visualization}, which are generated from single-camera high-resolution (3840x2160) front-view images of the vehicle. This image format can fully utilize the limited image resolution to provide a wide range of vehicle image information and improve the parallelism efficiency of the model. Furthermore, this work specifically designs a novel representation method for vehicles called Pseudo-3D Vehicle Representation (P3DVR). P3DVR consists of an extended BBox and a Side Projection Line (SPL), and its properties are derived from the projection of the vehicle 3D BBox onto the image plane based on the rigid body characteristics of the vehicle. This simplifies the post-inference process of the model prediction results. P3DVR can not only describe the position and size information but also describe the appearance and posture of the vehicle.
By incorporating geometric priors such as camera intrinsic parameters and ground plane, P3DVR can reconstruct the true 3D geometric structure of vehicles, providing important geometric cues for estimating the distance between the ego-vehicle and object vehicles\cite{1212895}, possible collision time\cite{1336352}, vehicle motion tasks\cite{8237543}, etc., without relying on expensive sensor data\cite{9153796}.
\IEEEpubidadjcol
\par
In this method, feature extraction and P3DVD are implemented by a fully convolutional network. 
The one-stage decoupled detection head\cite{ge2021yolox} is extended according to the composition of P3DVR. During training, in addition to independent attribute losses, a shape matching loss is introduced based on the joint constraint between expanded BBox and SPL. This shape matching loss provides holistic supervision information for model learning, enhancing the consistency of predicted results for extended BBox and SPL to describe the pseudo-3D vehicle and improving the stability and authenticity of P3DVR. Moreover, image augmentation based on window-following is applied to address the data pattern of DW images, ensuring that the Center Window (CW) images are always taken from the concentrated area of distant vehicles in original images, which helps the model better adapt to and utilize the image data in DW pattern during training. To evaluate the performance of P3DVD, we design corresponding evaluation metrics and the experimental evaluations demonstrate the excellent performance of our method in wide-range P3DVD.
\par
In summary, the main contributions of this work are as follows:
\begin{itemize}
    \item DW images is introduced, which supports wide-range vehicle detection with limited resolution. Window-following image augmentation method is designed correspondingly, which helps the model make full use of the DW image during training.
    \item P3DVR is specially designed for vehicles, which can provide not only the position and size of vehicles but also their pose information, as well as semantic information about their appearance.
    \item A joint constraint loss based on expanded BBox and SPL is proposed, which provides holistic supervision during training and improves the detection performance.
\end{itemize}

\section{Related work}
\subsection{Vehicle Detection}
In recent years, DNN-based object detection models have demonstrated remarkable performance in vehicle detection. Traditional two-stage detectors\cite{girshick2015fast, 7485869} employ region proposal methods\cite{uijlings2013selective, 7485869} to generate candidate BBox, while one-stage detectors like SSD\cite{liu2016ssd} and YOLO\cite{7780460} pre-define anchors in advance for efficient inference. However, these methods still have limitations, such as reliance on anchor and post-processing steps like NMS. Anchor-free methods like Fully FCOS\cite{tian2019fcos} directly predict BBox for each grid point on the feature map, providing flexibility for objects of different scales. Recently, DETR\cite{carion2020end}, an end-to-end object detection method, has gained attention for its set matching approach that eliminates NMS. The input of above detectors is the conventional front view image of vehicles, making it difficult to perform wide-range vehicle detection under the limitations of computational resources. Moreover, these detectors only provide axis-aligned BBox output, which lacks additional semantic and geometric information about the vehicles.
\par
Our proposed vehicle detector builds upon YOLOX\cite{ge2021yolox} framework and uses DW pattern images as input to achieve accurate wide-range vehicle detection with only a reasonably small increase in computational cost. On the other hand, our vehicle detector is a P3DVD detector, which can predict richer semantic and geometric information about the vehicle in addition to its position and size information in the image.
\subsection{Object Representation}
Recently,  various object representations that account for data and object characteristics have been proposed in addition to the common axis-aligned BBox. 2D segmentation methods\cite{8237584} classify pixels individually and output the external contours of objects. In the field of remote sensing, rotated BBox\cite{Liu2017LearningAR, yang2021r3det, 9272784, NEURIPS2021_98f13708} have been developed to accurately describe the rolling angle variation of objects. Additionally, inspired by text image detection, quadrilateral object representation methods\cite{9001201} have been proposed to better determine the object boundaries. Due to different application scenarios, these representations may not fully capture the shape and pose of vehicles. 
TBox\cite{9153796} extends from the axis-aligned BBox by adding additional information such as the visible surface category and three vehicle key points, but it requires a cascade-style approach and geometric reasoning.
\par
In this task, we represent the vehicle object as P3DVR. P3DVR consists of an extended BBox and SPL, which allow for the geometric decomposition of the forms that 3D BBox of vehicles projected onto the image based on the rigid body characteristics of vehicles. As a result, the attributes of P3DVR are fully decoupled and independently predicted, requiring only the combination of the predicted results of various attributes based on the Anchor\cite{7485869} to obtain the P3DVR representation.
\subsection{Shape Matching Constraint}
The shape of detected objects is represented by a composition of multiple attributes predicted by the network, 
and thus shape matching constraints can optimize the overall prediction performance of the model. Yang et al. \cite{NEURIPS2021_98f13708} transform rotated BBox's shape parameters into a 2D Gaussian distribution and compute the Kullback-Leibler Divergence (KLD) between Gaussian distributions as the regression loss. This allows the rotated BBox parameters to mutually influence each other adaptively and collaboratively during the dynamic joint optimization process. Liu et al.\cite{10030447} also convert the shape parameters of lane line segments into a 2D Gaussian distribution, and the KLD between Gaussian distributions is computed as an overall joint constraint loss.
\par
A joint constraint on the extended BBox and SPL provides a way to generate holistic supervisory information for multiple attributes of the joint target. It transforms all attributes of P3DVR into a unified 2D Gaussian distribution, which enables the model to establish intrinsic connections between the various attributes of P3DVR. This facilitates the model to adjust the predicted attributes in a coordinated manner during training, leading to more stable and accurate P3DVR predictions during testing.

\section{Method}
In this section, we first briefly propose our problem definition and then introduce our model architecture and some details during the training process. 
\subsection{Problem Definition}\label{Problem Definition}
According to the statistical analysis of the dataset, distant vehicles (vehicles' sizes are smaller than a predefined threshold in the image) are concentrated in the central region from the ego-vehicle's perspective. By extracting the central region as the CW image, which contains the raw information of distant vehicles, and obtaining the GW image by scaling the original image to provide contextual information about nearby vehicles. The DW images are created by vertically concatenating the CW and GW images. These DW images provide wide-range vehicle information for the model.
\begin{figure}
    \flushleft
    \subfloat[]{
    \label{fig:P3DVR}
    \hspace{-0.3cm}\includegraphics[width=0.6\linewidth]{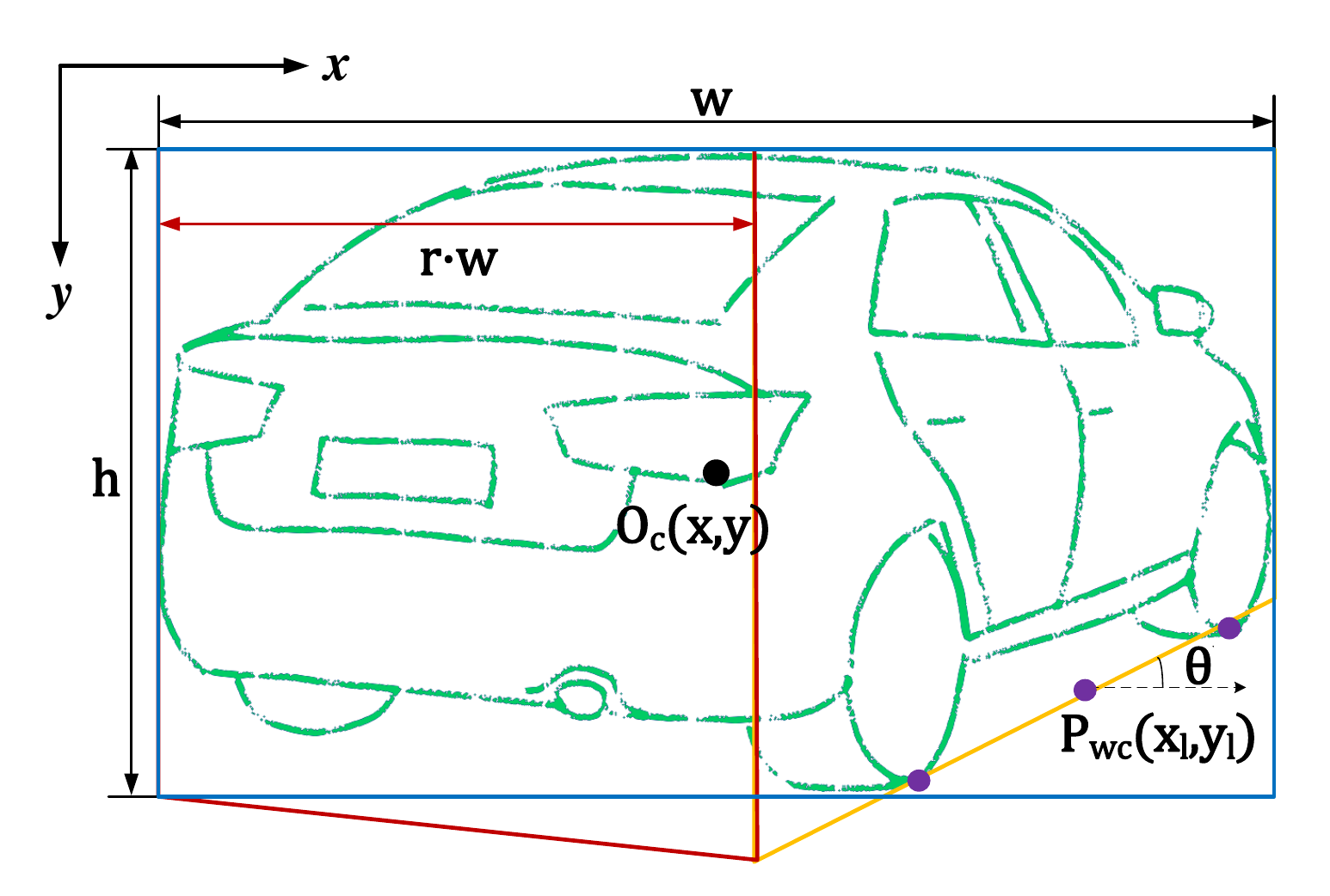}
    }
    \subfloat[]{
    \label{fig:pose_definition}
    \includegraphics[width=0.4\linewidth]{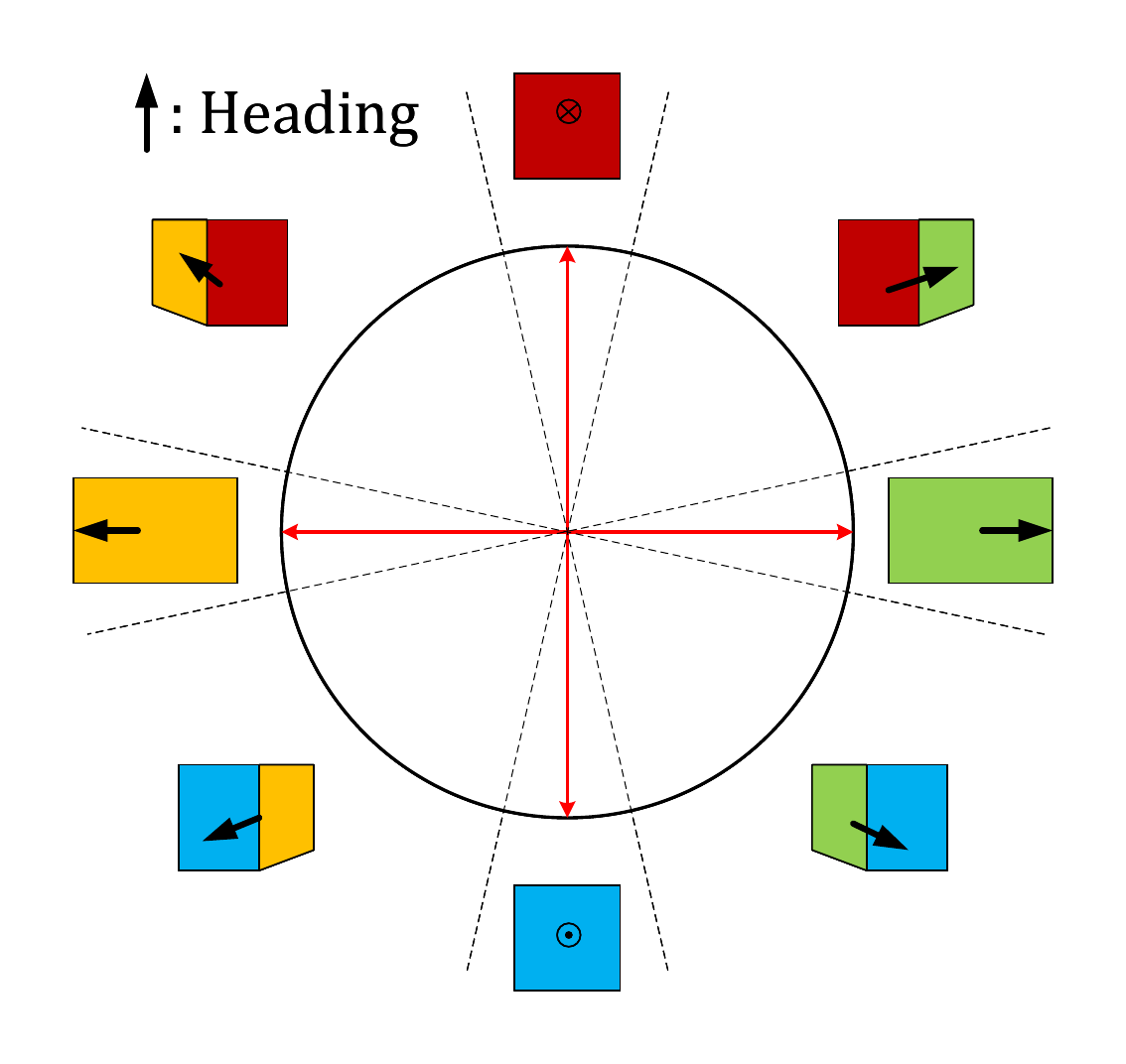}
    }
    \caption{Output representation. (a) Geometric attribute decomposition. (b) The pose is encoded 0-7 counterclockwise, and the arrow indicates the orientation of the vehicle head from the perspective of the camera}
    \label{fig:example_of_P3DVR}
\end{figure}
\subsubsection{Output Representation}\label{P3DVR}
The P3DVR we designed can represent the 3D BBox of a vehicle projected onto a 2D image. The rigid characteristics of the vehicle can be utilized to decompose the representation into geometric attributes, which are then combined to form an extended BBox and the SPL of the vehicle. The P3DVR can be uniquely determined by these two components, as illustrated in Fig. \ref{fig:P3DVR}.
\par
The extended BBox can be described by a set of shape parameters $EB = (x,y,w,h,r,p)$. In the figure, $O_c(x, y), w, h$ represent the center coordinates, width, and height of the BBox of the vehicle target in the image, respectively. In addition to these attributes, we extend the BBox with two additional properties based on the vehicle's pose in the camera field of view, namely $r$ representing the ratio of the width of the left part of the segmentation line to the total width of the BBox, and the pose classification result $p$ based on the orientation of the vehicle as shown in Fig. \ref{fig:pose_definition}. 
\par
The side projection line is the main form of the 3D geometric structure of the vehicle target, which can be described by a set of shape parameters $SPL = (p_{wc},\theta)$.  $p_{wc}$ is the midpoint of the line connecting the contact points of the front and rear wheels on the same side, and it can be easily described by its coordinate as $p_{wc} = (x_l,y_l)$. SPL is the connecting straight line past the front and rear wheel grounding points on the same side. According to the planar geometry prior in Fig. \ref{fig:P3DVR}, the SPL can be uniquely determined by midpoint $p_{wc}$ and the angle $\theta$ between this line and the x-axis of the image coordinate system. Therefore, the P3DVR can be represented as $(x,y,w,h,r,p, p_{wc},\theta)$.
\par
Given the P3DVR, the shapes of the red and yellow trapezoidal boxes in Fig. \ref{fig:example_of_P3DVR} can be determined, and these two trapezoidal boxes can describe some important 3D geometric structures of the vehicle.
\par
According to the composition of P3DVR, the P3DVD task can be decomposed into detecting the vehicle's BBox, pose classification, and ratio of the left part, and detecting SPL's middle point and its angle with the x-axis.

\begin{figure*}[tp]
    \centering
    \subfloat[]{
    \includegraphics[width=0.9\textwidth]{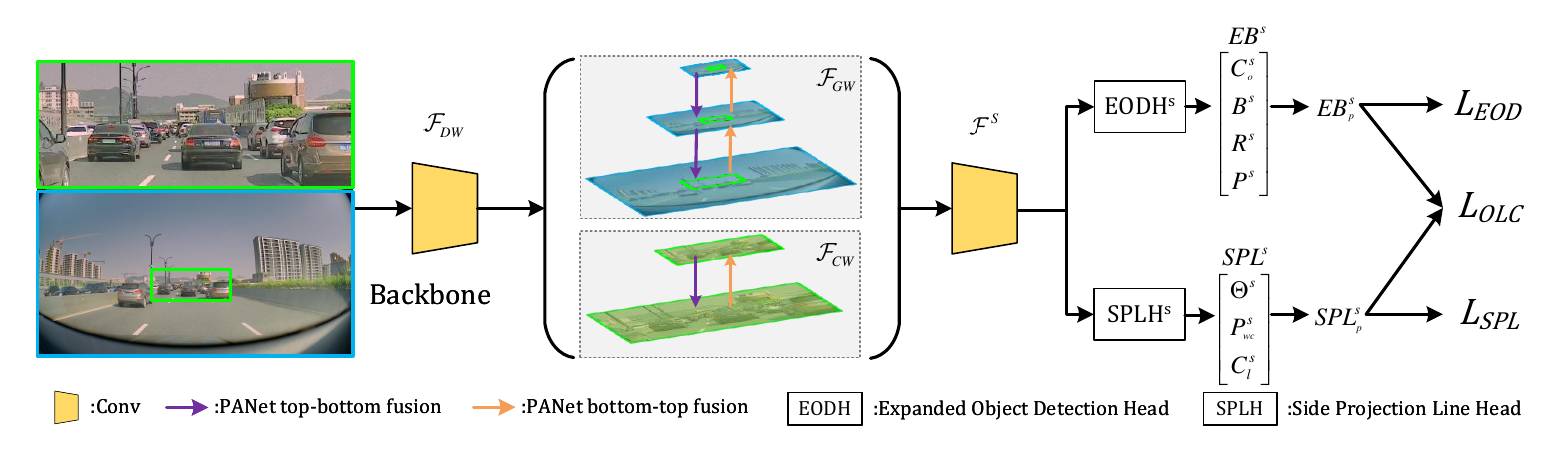}
    }\\
    \subfloat[]{
    \includegraphics[width=0.45\textwidth]{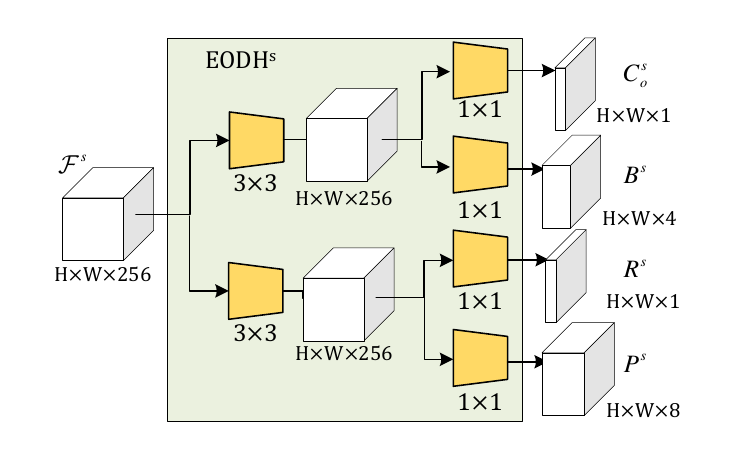}
    }
    \subfloat[]{
    \includegraphics[width=0.45\textwidth]{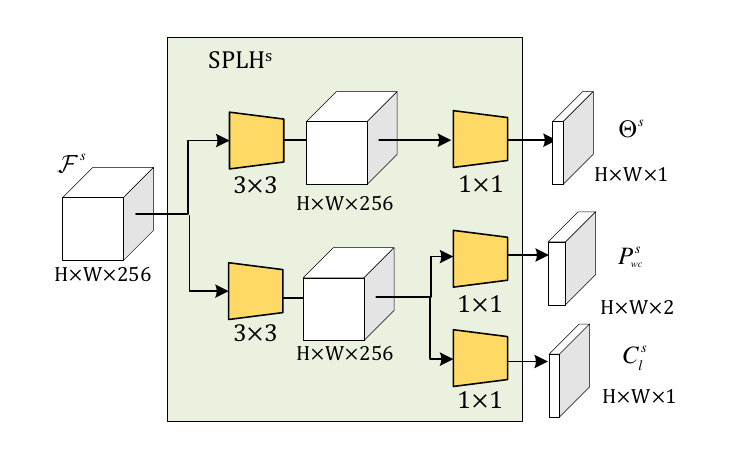}
    }
    \caption{
    P3DVD model architecture. (a) Brief network architecture. (b) Expanded Object Detection Head(EODH). (c) Side Projection Line Head(SPLH). The model takes DW images as inputs, and then the detection head module performs P3DVD on each scale of feature maps, generating predictions for various attributes of the P3DVR. 
    }
    \label{fig:network_architecture}
\end{figure*}

\subsection{Network Architecture}\label{Network Architecture}
Fig. \ref{fig:network_architecture} briefly shows the network architecture of our P3DVD. With respect to the P3DVR, we have designed corresponding pseudo-3D detection heads, including an Expanded Object Detection Head(EODH) and a Side Projection Line Head(SPLH).
\subsubsection{Backbone}
\par
The model employs CSPDarknet53\cite{redmon2018yolov3} as the backbone for feature extraction. As shown in the Fig. \ref{fig:network_architecture}a, the backbone directly extracts image features from synthesized DW images to improve computational efficiency, and outputs multi-scale features $\mathcal{F}_{DW}=\left\{F^{s_d}_{DW}|s_d=8,16,32\right\}$ from the DW images, where $s_d$ denotes the stride of $F^{s_d}_{DW}$ relative to the DW images. Due to the dual-scale characteristics of the DW images, $F^{s_d}_{DW}$ is composed of feature maps from two different scales. Therefore, during the multi-scale feature fusion process, it is necessary to first separate $\mathcal{F}_{DW}$ based on concatenation information, and then effectively fuse\cite{8579011} the feature map information between adjacent scales according to the scale information. The separation of $\mathcal{F}_{DW}$ results in GW multi-scale features denoted as $\mathcal{F}_{GW}=\left\{F^{s_g}_g|s_g=8s_o,16s_o,32s_o\right\}$, and CW multi-scale features denoted as $\mathcal{F}_{CW}=\left\{F^{s_c}_c|s_c=8,16\right\}$, where $s_g$ and $s_c$ represent the stride of $F^{s_g}_g$ and $F^{s_c}_c$ with respect to the original image respectively, and $s_o$ represents the scaling factor of GW image with respect to the original image.
\subsubsection{EODH}
As shown in Fig. \ref{fig:network_architecture}b, the EODH is responsible for predicting 2D object properties based on the extended vehicle pose. In addition to outputting the BBox $B \in \mathbb{R}^4$ and the objectness confidence $C_o$, EODH also predicts the regression value $R$ for the width ratio of the left part of the vehicle segmentation line, as well as the pose classification scores $P \in \mathbb{R}^8$ for the vehicle's pose on the ground plane from the camera's viewpoint. The prediction output of EODH can be represented as follows:
\begin{equation}
    EB^s = \left[C^s_o, B^s, R^s, P^s\right] = EODH^s(F^s)
    \label{EODH_expression}
\end{equation}
where $s$ represents five different strides of the feature $F^s$ composed of $F^{s_g}_g$ and $F^{s_c}_c$.
\subsubsection{SPLH}
As illustrated in Fig. \ref{fig:network_architecture}c, the SPLH is responsible for predicting the SPL properties of pseudo-3D vehicles. According to the definition of SPL in Section \ref{P3DVR}, SPLH needs to output the normalized value of the SPL angle $\Theta$, as well as the coordinates of the midpoint $P_{wc}$ of the line connecting the contact points of the same side front and rear wheels of the vehicle. It is worth noting that the point coordinate prediction by SPLH heavily relies on the visible image information of the wheels on the same side of the vehicle. However, when the front and rear wheels of some vehicles are not visible or occluded in the image, the model cannot explicitly obtain the wheel-ground contact information from the image and needs to rely on long-term training to form empirical knowledge to predict the contact point coordinates from other features, which may result in unstable P3DVR prediction results. Therefore, a confidence measure $C_l$ is added to determine the prediction results of the current point coordinates in SPLH. Although the definition of the SPL angle in Section \ref{P3DVR} also depends on the information of the contact points of the same side front and rear wheels, based on the rigid body characteristics of vehicles, it can be assumed that the angle of the SPL is equivalent to the inclination angle of the vehicle body in the image coordinate system. Therefore, SPLH can predict the SPL angle from the overall features of the vehicle, without overly relying on the information of the contact points of the same side front and rear wheels. The prediction output of SPLH can be represented as follows:
\begin{equation}
    SPL^s = \left[\Theta^s, P^s_{wc}, C^s_l\right] = SPLH^s(F^s)
    \label{SPLH_expression}
\end{equation}

\subsection{Loss Function}
Based on the output content of EDOH and SPLH, this section introduces the design of loss function for the P3DVD model. For N positive target samples assigned by the network in a batch, the average loss is calculated for the predicted results $\left\{(EB^s_i, SPL^s_i)|i=1, 2, ..., N\right\}$. It is worth noting that, all loss calculations are performed in the pixel coordinate system scale, so the prediction outputs $EB^s$ and $SPL^s$ in the feature coordinate system need to be transformed to $EB^s_p$ and $SPL^s_p$ in the pixel coordinate system according to the corresponding scale factors $s$. As shown in Fig. \ref{fig:network_architecture}, the overall training loss consists of the Expanded Object Detection Loss $L_{EOD}$, the Side Projection Line Detection Loss $L_{SPL}$, and the Shape Matching Loss based on the joint constraint between the EB and SPL $L_{OLC}$, formulated as follows:
\begin{equation}
    L = L_{EOD} + L_{SPL} + L_{OLC}
    \label{overrall_loss}
\end{equation}


\subsubsection{Expanded Object Detection Loss}
$L_{EOD}$ is composed of the vehicle object confidence loss $L_O$, the BBox IOU loss $L_{IOU}$, the regression loss for the occupancy ratio on the left side of the vehicle segmentation line $L_R$, and the pose classification loss $L_{Pose}$, formulated as follows:
\begin{equation}
    L_{EOD} = L_O + \alpha_1L_{IOU} + \alpha_2 L_R + L_{Pose}
    \label{L_EOD}
\end{equation}
The weight coefficients for the losses are denoted as $\alpha_1$ and $\alpha_2$ in the equations. $L_O$ and $L_{Pose}$ are calculated using binary cross-entropy. $L_{IOU}$ is calculated based on the definition provided in reference \cite{10.1145/2964284.2967274}. $L_R$ is calculated using the L1 loss.

\subsubsection{Side Projection Line Detection Loss}
$L_{SPL}$ consists of the angle regression loss $L_\Theta$, point regression loss  $L_{PC}$, and the confidence loss $L_{Conf}$ for the midpoint prediction, formulated as follows:
\begin{equation}
    L_{SPL} = \alpha_3 L_\Theta +L_{Conf} + \alpha_4 L_{PC}
    \label{L_SPL}
\end{equation}
\begin{figure}[htbp]
    \centering
    \includegraphics[width=0.85\linewidth]{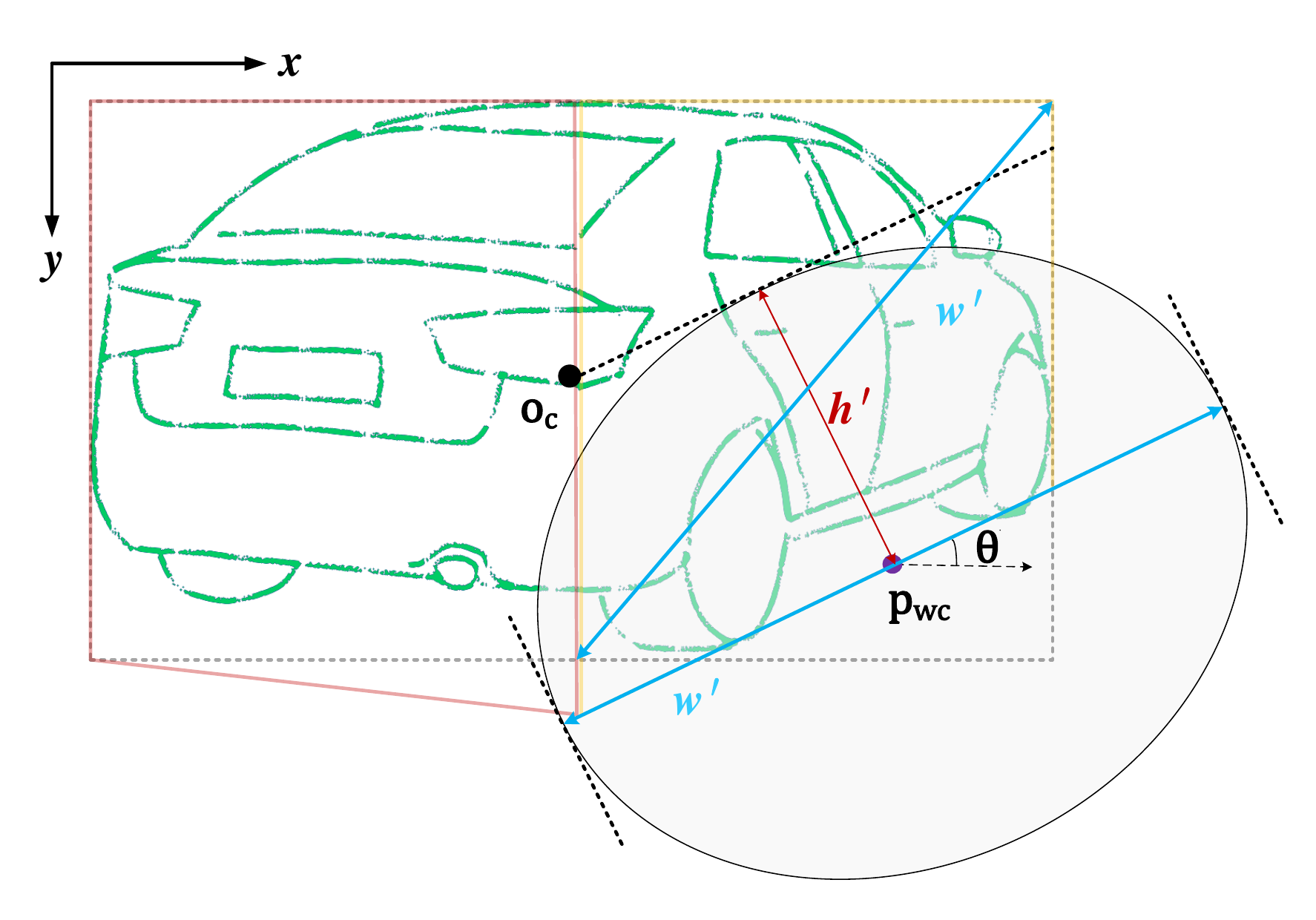}
    \caption{Geometric relationship between P3DVR and 2D Gaussian distribution. The mean of the 2D Gaussian distribution is the midpoint $p_{wc}$ of the line connecting the front and rear wheels of the vehicle on the same side. The variance in the $y$ direction is the distance $h'$ from the center point $O_c$ of the BBox of P3DVR to the SPL, and the variance in the $x$ direction is $\frac{w'}{2}$. The rotation angle is the inclination angle $\theta$ of the SPL.}
    \label{fig:Geometric Relationship between PSV3D targets and two-dimensional Gaussian distributions}
\end{figure}
In the formula, $\alpha_3, \alpha_4$ represent the loss weight. According to the definition of $conf$, its supervision information is generated based on whether the grounding points of the front and rear wheel on the same side of the vehicle’s ground truth label are present or not. The supervision label for $conf$ is set to 1 if the grounding points are present, and 0 if they are not. Therefore, the loss for the predicted results of $conf$ can be calculated using binary cross-entropy. The output $p_{wc}$ from SPLH represents the predicted regression of the point relative to the anchor center, and the predicted angle $\theta$ is normalized to the range of [-1, 1]. Then, the L1 loss can be used to calculate the training loss for these two predictions. 

\subsubsection{Joint Constraints loss of EB and SPL}
To compute $L_{OLC}$, we first convert the P3DVR into a 2D Gaussian distribution \cite{NEURIPS2021_98f13708}. The shape matching loss is then calculated as the KLD between these Gaussian distributions. The conversion process from P3DVR to 2D Gaussian distributions is exemplified in Fig. \ref{fig:Geometric Relationship between PSV3D targets and two-dimensional Gaussian distributions}, the $i$-th target with SPL can be described as a 2D Gaussian distribution $G^i_p(\mu,\sum)$, and the mean $\mu$ and variance $\sum$ can be represented as follows:
\begin{equation}
    \begin{aligned}
        \mu &=p_{wc} \\
        \sum &=
        \begin{pmatrix}
            \cos\theta & -\sin\theta \\
            \sin\theta & \cos\theta
        \end{pmatrix}
        \begin{pmatrix}
            \frac{w'}{2} & 0 \\
            0            & h'
        \end{pmatrix}
        \begin{pmatrix}
            \cos\theta  & \sin\theta \\
            -\sin\theta & \cos\theta
        \end{pmatrix}
    \end{aligned}
    \label{Formula for calculating mean and variance of two-dimensional Gaussian distribution}
\end{equation}
where $h'$ is the distance from the center of BBox $O_c$ to SPL, and $w'$ is the diagonal length of the rectangular box on the side of the vehicle (the side that can see both front and rear tires from the camera’s perspective).

\par
Due to the asymmetry of KL divergence, this method uses bidirectional KLD proposed by Wu et al\cite{wu2021r} for calculation:
\begin{equation}
    D^i_{KL} = \frac{1}{2}[D_{KL}(G^i_p||G^i_t) + D_{KL}(G^i_t||G^i_p)]
    \label{D_KL calculation}
\end{equation}
where $G^i_p$ and $G^i_t$ respectively represent the 2D Gaussian distributions inferred from the predicted results and the ground truth of the $i$-th object. 
\par 
Finally, the SPL shape matching loss is constrained by the object and SPL, and can be shown as follows:
\begin{equation}
    L_{OLC} = \frac{\alpha_5}{N'} \sum_{i=1}^{N'} \left(1 - \frac{1}{1+\ln(D^i_{KL} + 1)}\right)
    \label{L_kl loss}
\end{equation}
\begin{figure}
    \centering
    \subfloat[]{
        \label{fig:non-enhance}
        \includegraphics[width=0.45\linewidth]{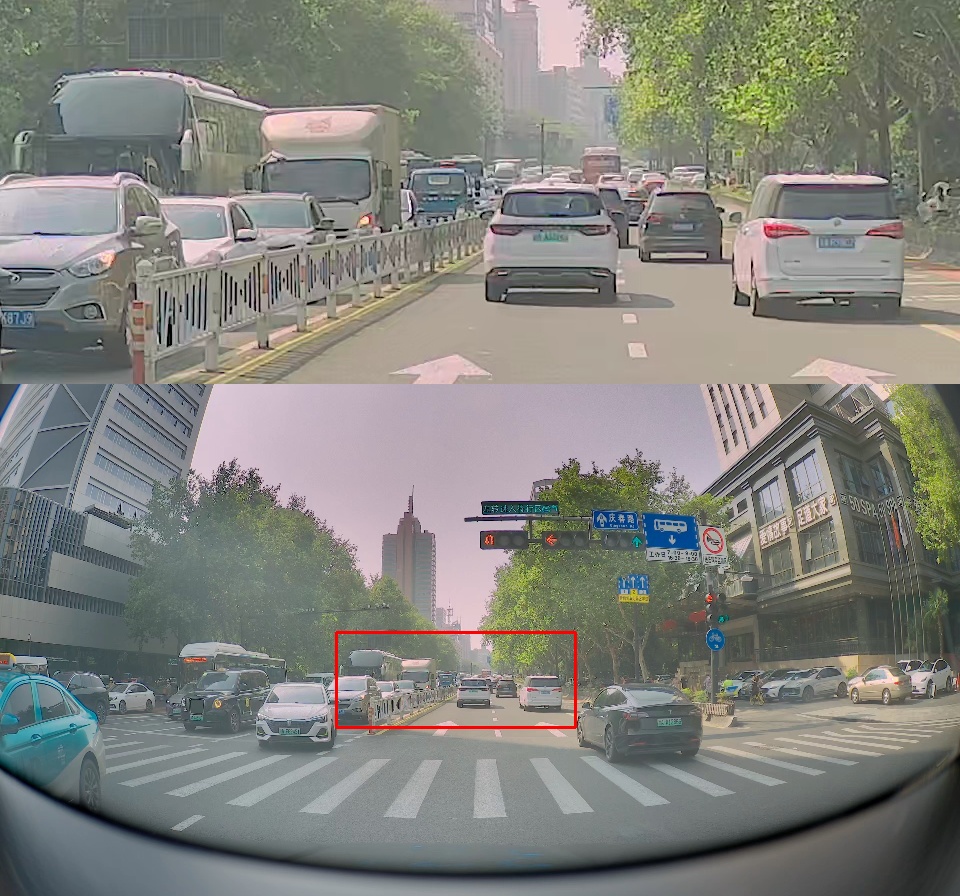}
        }
    \subfloat[]{
        \label{fig:mosaic+affine transformation}
        \includegraphics[width=0.45\linewidth]{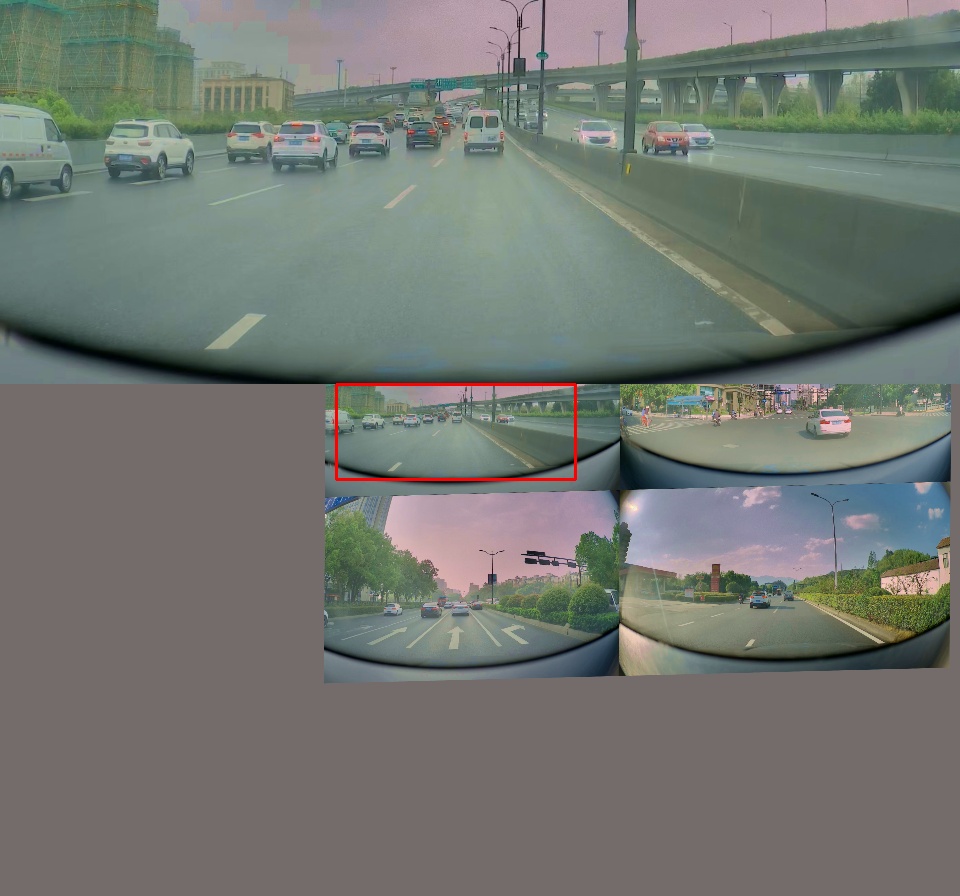}
    }
    \caption{DW images data processing results. The CW image is above the DW image, and the GW image is below it. The CW image is the content of the red rectangle area of the GW image in the original image. (a) shows the processing result without any image augmentation, and the window center position is taken from the predefined coordinates; (b) shows the enhancement result with mosaic and affine transformation, due to the transformation following the window center coordinates, the CW image can still be taken from the concentrated area of distance vehicles in one of the original images participating in mosaic enhancement.}
    \label{fig:Double-Window Image Data Processing Results}
\end{figure}
where $\alpha_5$ represents the loss weight, $N'$ represents the number of targets with SPL.
\par
According to the expression form of the vehicle 3D BBox projected onto the 2D image, the vehicle SPL is the main form to present the 3D geometric structure of vehicles in images. Given that P3DVR is sensitive to the slope and position of SPL, even slight differences between the predicted SPL and the ground truth can lead to a lack of authenticity in the predicted P3DVR. To address this issue, $L_{OLC}$ is added based on the geometric constraint relationship between P3DVR attributes during the training stage. This approach introduces holistic supervision information for model learning, improving the stability and authenticity of P3DVR predicted by the model. 


\subsection{Window-Following Image Augmentation}\label{img_aug}

Image augmentation is crucial for model training, but existing methods cannot fully leverage the advantages of DW images due to their unique structure. According to the DW images synthesis method in Section \ref{fig:DW_Visualization}, the CW must be taken from the concentrated area of distance vehicles in the original image. However, most data augmentation techniques alter the structure of the original image, which can shift the position of the concentrated area of distance vehicles. If we still obtain the CW from the predefined position, it will be difficult to ensure that it covers the concentrated area of distance vehicles, leading to a lack of fine-grained information for the model. To solve this issue and ensure that the CW images are always taken from the concentrated area of distance vehicles in the original image, we introduce a window-following image augmentation method. This method performs the same transformation as the image pixel coordinates on the predefined CW center position during the data enhancement process. After all augmentation operations are completed, CW is then taken according to the center coordinate position of the windows that follows the transformation. As shown in Fig. \ref{fig:Double-Window Image Data Processing Results}, the window-following image augmentation guarantees that CW images are always obtained from the concentrated areas of distance vehicles, regardless of geometric transformations during image augmentation.

\begin{figure*}[htbp]
    \centering
    \subfloat[]{
    \includegraphics[width=0.4\textwidth]{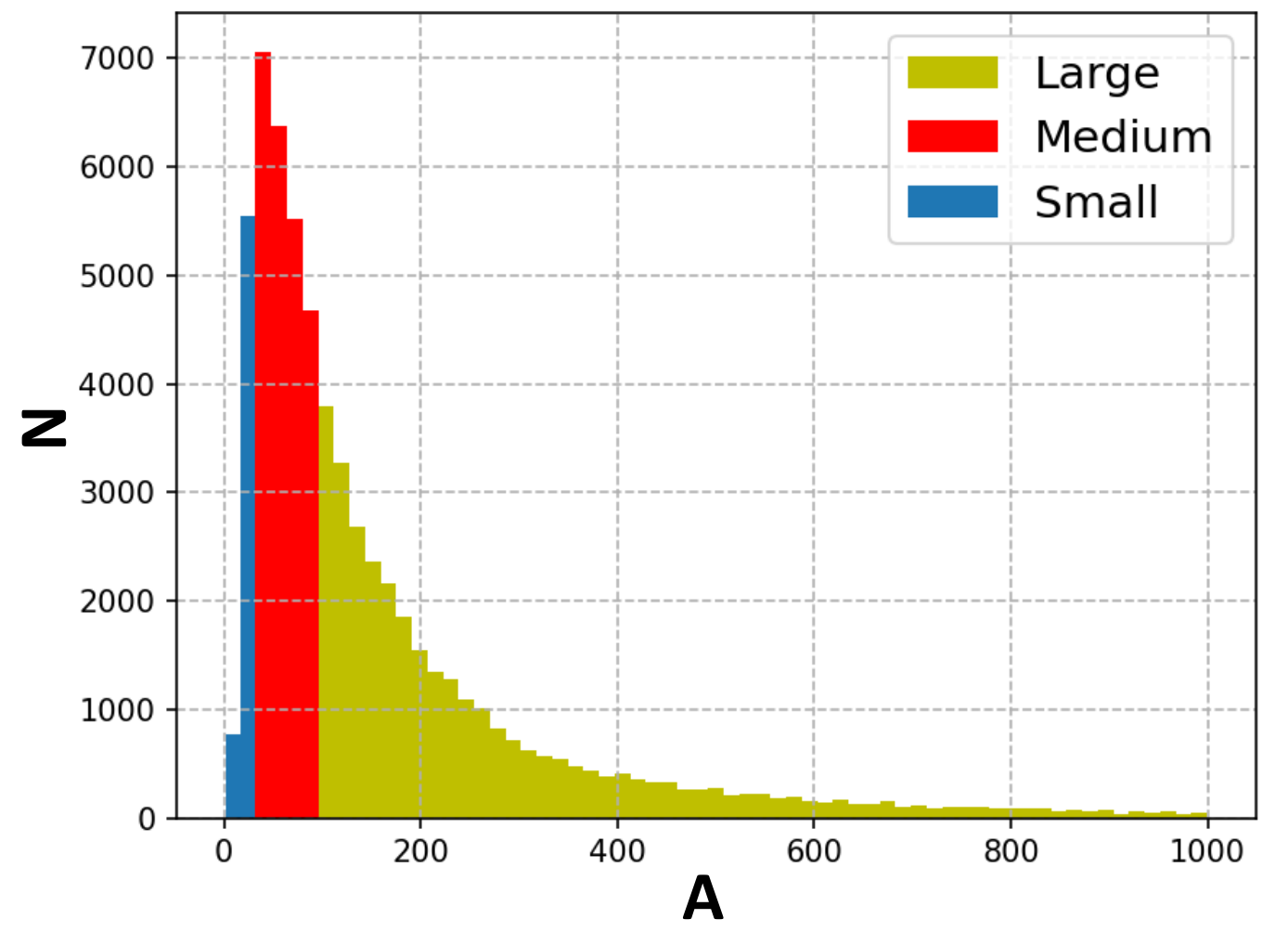}
    }
    \subfloat[]{
    \includegraphics[width=0.4\textwidth]{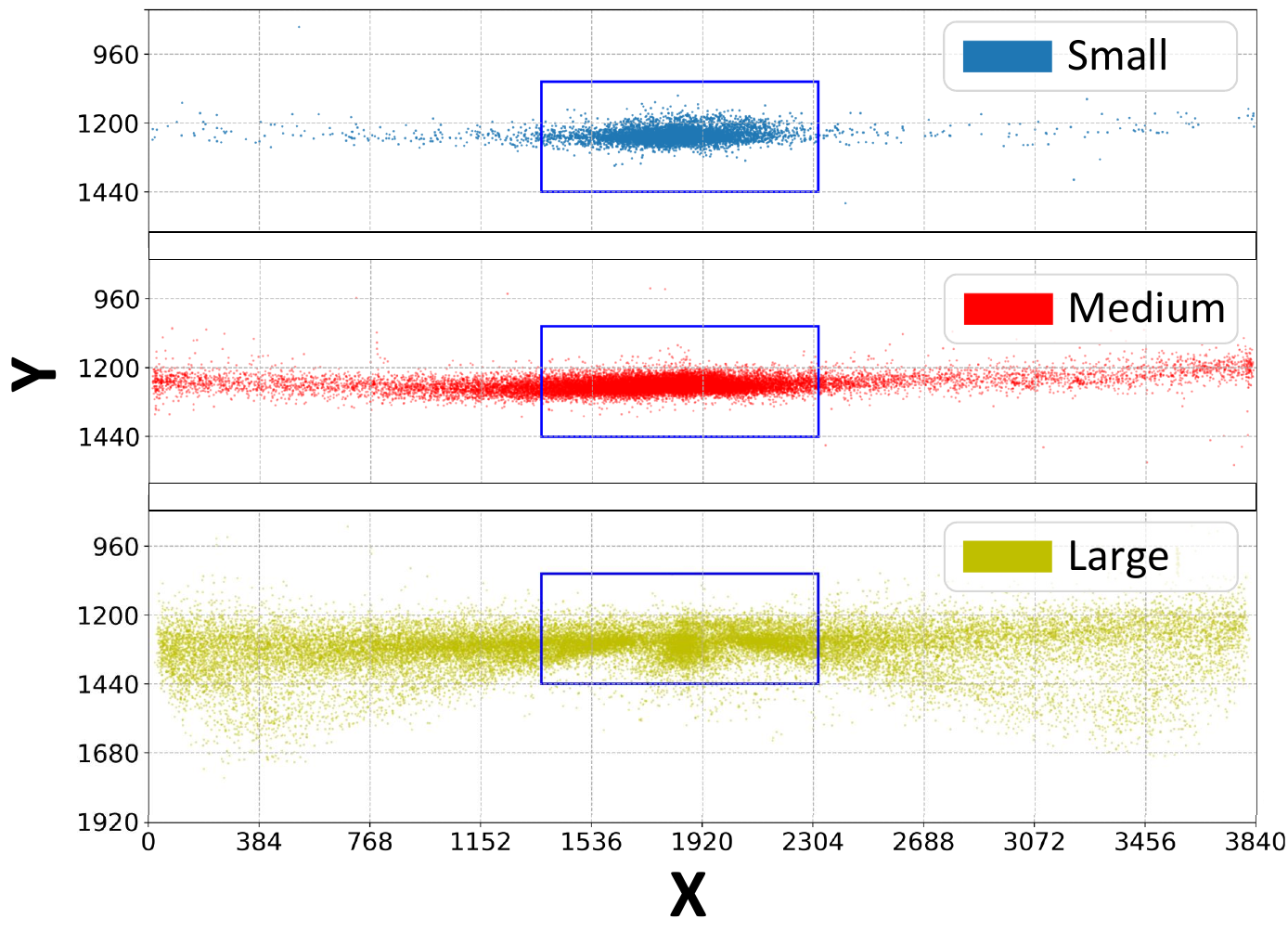}
    }
    \caption{The statistical histograms and location distribution scatter plots of objects with different sizes in the training dataset. The objects are divided by their area into small: (0, 1024], medium: (1024, 9216], large: (9216, +$\infty$). (a) Statistical histograms. The A on the horizontal axis represents the arithmetic square root of the labeled box area, and the N on the vertical axis represents the number of boxes within that area range. (b) Object location distribution scatter plots. X and Y represent the pixel coordinate value of the input image in the training dataset.}
    \label{fig:dataset}
\end{figure*}

\section{Experiment}
This section presents various comparative and ablation experiments to demonstrate the feasibility and superiority of the proposed wide-range P3DVD method based on DW images.

\subsection{Dataset}
The dataset used in this study consists of images captured by a front-view camera mounted on a vehicle, which were taken in various traffic scenarios on urban and highway roads, including both daytime and nighttime, clear and rainy weather conditions. The images have a resolution of 3840x2160 and the Horizontal FOV is $120^{\circ}$ and Vertical FOV is $54.8^{\circ}$. The dataset provides BBox annotations for all vehicles in the images, as well as the segmentation ratio of the left part, and the orientation pose of the vehicles on the ground plane under the camera's perspective. Only the vehicles with visible front and rear wheels were annotated with wheel grounding points information. For experiments, a total of 5964 images are randomly selected from the dataset for training, and 3376 images are used for validation.
\par
In Fig.\ref{fig:dataset}a, the statistical histogram shows that large objects account for the majority (52.63$\%$) of annotations, while medium and small objects have a lower proportion (37.37$\%$ and 10.00$\%$). In the training dataset, the largest box has an area of $4.01\times10^6$ square pixel, the minimum is 2.94 square pixel, and the average area of the annotation box is $2.93\times10^4$ square pixel. On average, each image has 10.6 targets. The object location distribution plots in Fig.\ref{fig:dataset}b illustrate the distribution position of large, medium, and small targets in the images of training dataset. It can be observed that the distribution of large and medium targets is relatively scattered, while the distribution of small targets is relatively concentrated, which verifies the hypothesis that distant and small targets are concentrated in the Center Window proposed earlier. According to statistics, the coordinates of the CW's midpoint is (1840, 1248).

\subsection{Experiment Settings}
\subsubsection{Evaluation Metrics}\label{Evaluation Metrics}
The commonly used evaluation method\cite{lin2014microsoft} for image-based 2D object detection tasks only assesses the prediction of object categories and bounding boxes (BBox) and is unable to evaluate other attributes of P3DVR. To address this, we propose an evaluation method for P3DVD that first computes the BBox IOU between the prediction and the ground truth, and considers a target a true positive when the IOU exceeds a threshold $\lambda$. We then calculate the attribute prediction error against the ground truth and determine if the prediction is correct based on an error threshold $\lambda'$.
Finally, we compute the ratio of correctly detected attributes to the total number of true positives as the average precision (AP) of the attribute under the threshold values $\lambda$ and $\lambda'$. The evaluation of each P3DVR attribute is expressed as follows:
\begin{equation}
    \begin{aligned}
        &\mathbb{T} = \left\{ \left( i, j \right) | \mbox{IOU}(\hat{B}_i, B_j) \geq \lambda, i \in [0, N_g), j \in [0, N_p)  \right\} \\
        &\mbox{ABP}@\lambda @\lambda'_b = \frac{1}{N_T} \sum_{(i, j) \in \mathbb{T}} \delta(1 - \mbox{IOU}(\hat{B}_i, B_j)\leq\lambda'_b)\\
        &\mbox{ARP}@\lambda @\lambda'_r = \frac{1}{N_T} \sum_{(i, j) \in \mathbb{T}} \delta(|\hat{r}_i - r_j|\leq\lambda'_r) \\
        &\mbox{PP}@\lambda = \frac{1}{N_T} \sum_{(i, j) \in \mathbb{T}} \delta\left(\hat{pose}_i=pose_j \right) \\
        &\mbox{AAP}@\lambda @\lambda'_a = \frac{1}{N'_T} \sum_{(i, j) \in \mathbb{T}'} \delta(|\hat{\theta}_{ni} - \theta_{nj}| \leq\lambda'_a) \\
        &\mbox{APP}@\lambda @\lambda'_p = \frac{1}{N'_T} \sum_{(i, j) \in \mathbb{T}'} \delta(||\hat{p}_{wci} - p_{wcj}||\leq\lambda'_p)
    \end{aligned}
    \label{Evaluation_formulation}
\end{equation}
\begin{table*}[htbp]
    \centering
    \caption{Comparative experiment on input image pattern. The objects are evaluated based on their area, divided into small, medium, large, and all, following the evaluation method for P3DVD proposed in Equation.\ref{Evaluation_formulation}. All the metrics are the average precision of prediction evaluation results under different IOU and attribute error thresholds which are defined in Section.\ref{Evaluation Metrics}. The metrics are taken as a percentage.}
    \label{tab:comparative experiment on input image pattern}
    \resizebox{\textwidth}{!}{
        \begin{tabular}{lcccccccccccc}
            \toprule
            Methods & Input & Area & ABP & ARP & PP & AAP & APP & AP & AR & Score & Parameters(M) & FLOPs(G)  \\ \midrule
            \multirow{4}{6em}{Baseline} & \multirow{4}{*}{$\mbox{GW}'$} & small & 28.67 & 61.16 & 55.88 & 68.18 & 40.84 & 1.87 & 2.09 & 36.96 & \multirow{4}{*}{0.93} & \multirow{4}{*}{6.02}  \\ 
            &     & medium & 39.03 & 60.92 & 68.97 & 73.08 & 73.82 & 23.99 & 28.12 & 52.56 \\
            &     & large  & 59.90 & 72.05 & 80.01 & 84.58 & 68.63 & 61.03 & 64.93 & 70.16 \\
            &     & all    & 53.49 & 69.15 & 77.09 & 82.88 & 69.35 & 37.55 & 40.67 & 61.46 \\ 
            \specialrule{0em}{2.5pt}{2.5pt}
            \Xhline{0.1pt}
            \specialrule{0em}{2.5pt}{2.5pt}    
            \multirow{4}{6em}{Ours} & \multirow{4}{*}{DW} & small & +10.81 & +18.78 & +21.56 & +6.52 & +48.25 & +30.86 & +38.66 & +25.06 & \multirow{4}{*}{+0.14} & \multirow{4}{*}{-0.64}  \\ 
            &     & medium & +10.06 & +5.72 & +6.12 & +4.52 & +11.52 & +20.04 & +20.05 & +11.23 \\
            &     & large  & -0.28 & -0.67 & +0.26 & +0.98 & +0.56 & +1.68 & +1.33 & +0.55 \\
            &     & all    & +0.38 & +1.52 & +1.22 & +0.95 & +30.62 & +13.51 & +14.40 & +5.08 \\ \bottomrule
        \end{tabular}
    }
\end{table*}
\begin{figure*}[htbp]
    \centering
    \subfloat[]{
    \includegraphics[width=0.31\textwidth]{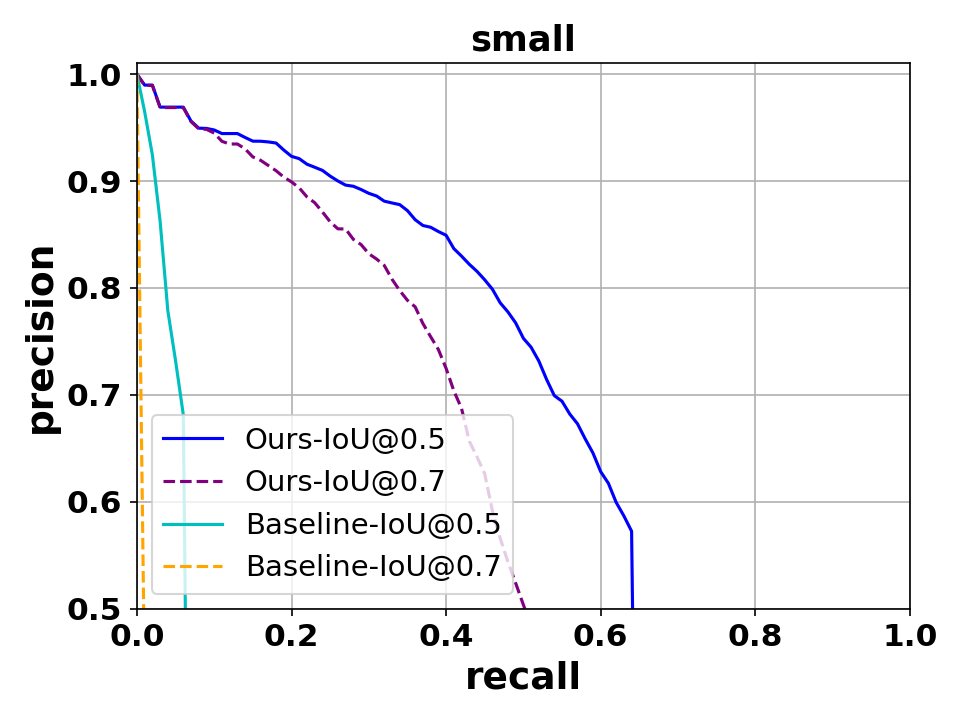}
    }
    \subfloat[]{
    \includegraphics[width=0.31\textwidth]{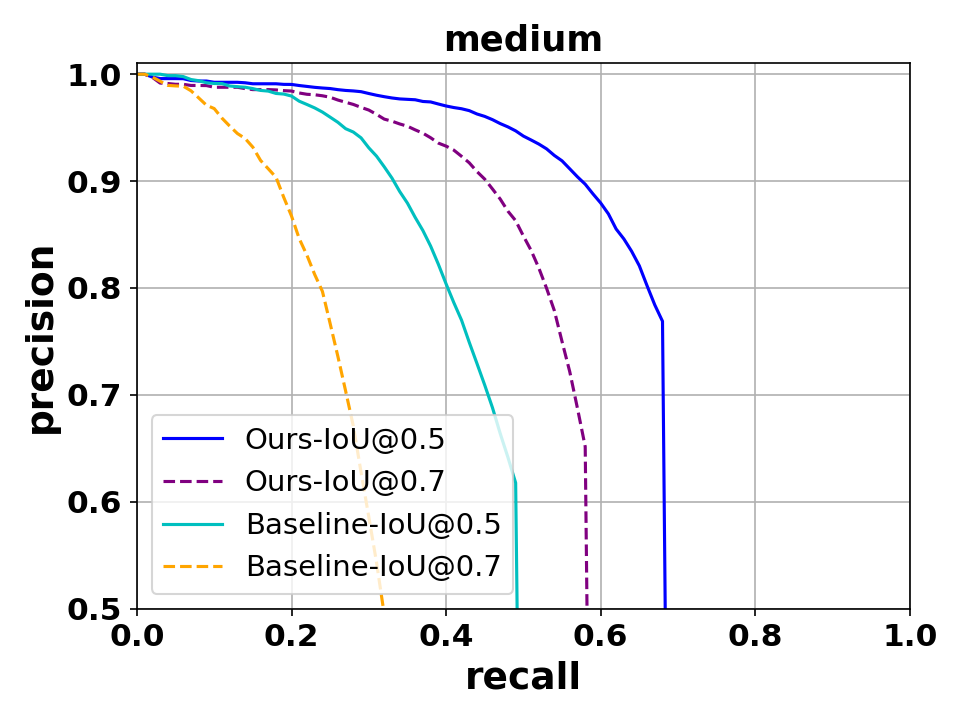}
    }
    \subfloat[]{
    \includegraphics[width=0.31\textwidth]{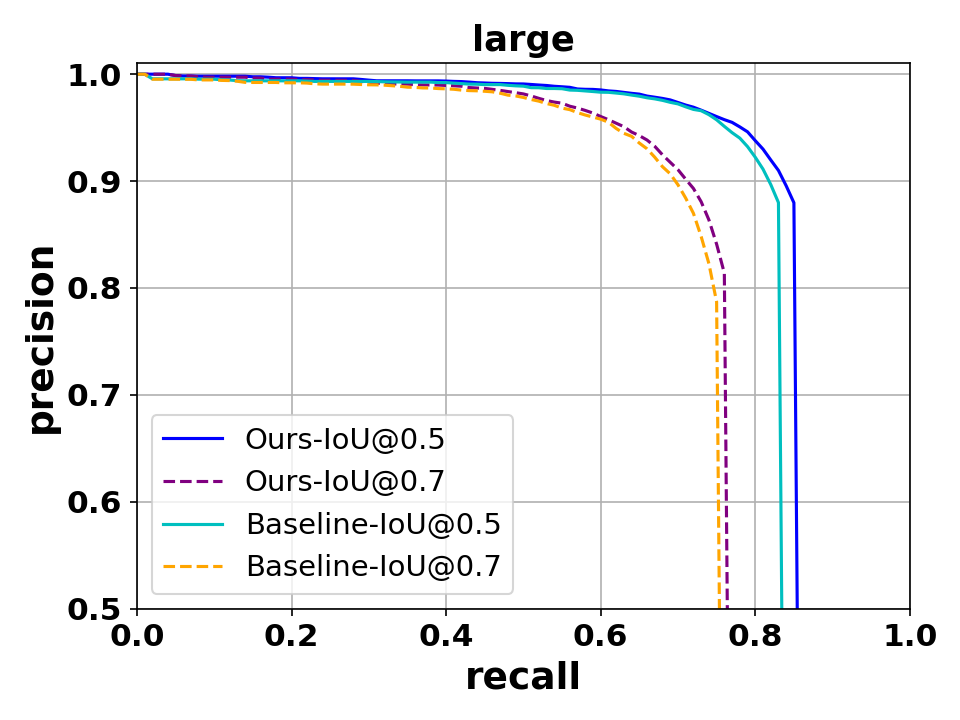}
    }
    \caption{The detecting performance of ours and the baseline method in different object sizes. (a) Small size. (b) Medium size. (c) Large size. }
    \label{fig:P-R}
\end{figure*}
In the Equation.\ref{Evaluation_formulation}, $ABP@\lambda@\lambda'_b$, $ARP@\lambda@\lambda'_r$, $PP@\lambda$, $AAP@\lambda@\lambda'_a$, and $APP@\lambda@\lambda'_p$ are defined as the precision of prediction results under different IOU and attribute error thresholds, respectively for BBox, vehicle segmentation ratio, pose classification, SPL angle, and the mid-point of the front and rear wheel grounding points. "@" denotes the evaluation result of each metric at a certain threshold, while $\lambda'_b$, $\lambda'_r$, $\lambda'_a$, and $\lambda'_p$ are the corresponding attribute prediction error thresholds. $N_g$ and $N_p$ are the number of ground truth and predicted objects, respectively, $N_T$ is the number of true positives, $\delta$ represents the comparison function, $\mathbb{T}'$ denotes the matching set of predicted objects that have ground truth object with wheel grounding point labels, and $N'_T$ is the number of objects in $\mathbb{T}'$. 
\par
To evaluate overall performance, ABP, ARP, PP, AAP, APP are used to represent the average precision of $ABP@\lambda@\lambda'_b$, $ARP@\lambda@\lambda'_r$, $PP@\lambda$, $AAP@\lambda@\lambda'_a$, and $APP@\lambda@\lambda'_p$. Specifically, $\lambda$ is selected from 0.5:0.05:0.95 (an arithmetic sequence from 0.5 to 0.95 with a common difference of 0.05), $\lambda'_b$ from 0.02:0.02:0.2, $\lambda'_r$ from 0.01:0.01:0.1, $\lambda'_a$ from 0.01:0.01:0.1, and $\lambda'_p$ from 2:2:20.
\par
In addition, the $AP@\lambda$ and $AR@\lambda$ metrics\cite{lin2014microsoft} of object detection are also important evaluation metrics. To evaluate the overall detection performance of the P3DVD model, the average AP and AR are also taken as the final model score.

\subsubsection{Implementation Details}
In the training stage, we adopted a dynamic-k assign strategy with an expandable cost term, similar to simOTA\cite{ge2021yolox}. This strategy allows using the predicted results from all five feature maps together for the assignment of positive and negative samples and the loss calculation. Correspondingly, in the inference stage, we perform non-maximum suppression (NMS) on the predictions from all five scales to obtain the final detection results.
\par
To reduce computational costs, this work constructs a lightweight network using depthwise separable convolutions \cite{8099678, AndrewHoward2017MobileNetsEC} and channel compression. The entire model is trained for 200 epochs in mixed precision mode with 64 images per batch. The augmentation techniques used during training include image translation, shear transformation, scaling, horizontal flipping, and mosaic augmentation. The window-following step described in Section \ref{img_aug} was also used in conjunction with the aforementioned augmentation techniques. After all data augmentation, a 960x384-sized image is extracted from the image as a CW image based on the window center coordinate position. Then, 52 and 60 rows are cropped from the top and bottom of the image, respectively, to obtain a 3840x2048-sized image, which is then scaled down to 1/4 to obtain a 960x512-sized GW image. Finally, the CW and GW images are concatenated vertically to form a 960x896 DW image as input to the model. The model is trained using the SGD optimizer, with an initial learning rate dynamically adjusted based on the batch size: $0.01 \times \mbox{batch size} / 64$. The learning rate is updated using the cosine annealing strategy during training. In the testing stage, the NMS post-processing operation is performed with a predicted target confidence and IOU threshold of 0.5 and 0.65, respectively. Both training and testing are completed on a single Nvidia RTX3090 GPU. 
\subsection{Comparative Experiment}
\begin{table*}[!t]
    \caption{Ablation Experiment: The effect of adding window-following(W-F) image augmentation and $L_{OLC}$ to the model training stage on detection performance. All metrics are reported in percentages. The first row of data in the table is the evaluation result of the control group, and the second and third rows of data benchmarks are the first and second rows respectively.}
    \centering
    \label{tab:Ablation Experiment}
    \begin{tabular}{cccccccccc}
        \toprule
        W-F & $L_{OLC}$ & ABP   & ARP   & PP    & AAP   & APP   & AP    & AR    & Score \\ \midrule
                         &           & 51.23 & 69.99 & 78.30 & 83.16 & 71.13 & 47.21 & 51.69 & 64.68\\
        \Checkmark       &           & +2.95 & +0.83 & +0.41 & +0.51 & +1.91 & +3.88 & +3.49 & +1.99\\
        \Checkmark       & \Checkmark& -0.32 & -0.16 & -0.42 & +0.16 & -0.07 & -0.03 & -0.12 & -0.14 \\ \bottomrule
    \end{tabular}
\end{table*}
\begin{figure*}[htbp]
    \centering
    \subfloat[]{
        \includegraphics[width=0.22\textwidth]{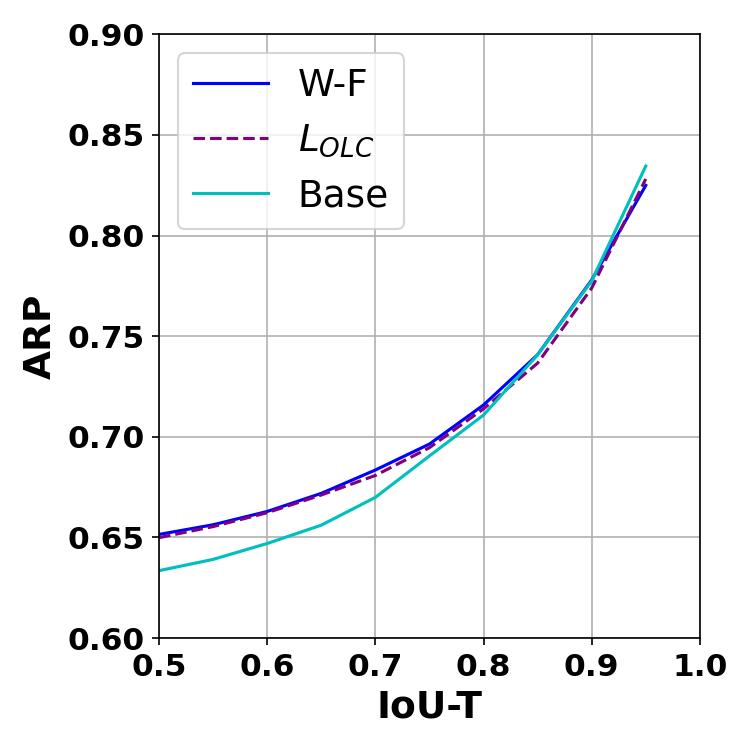}
    }
    \subfloat[]{
        \includegraphics[width=0.22\textwidth]{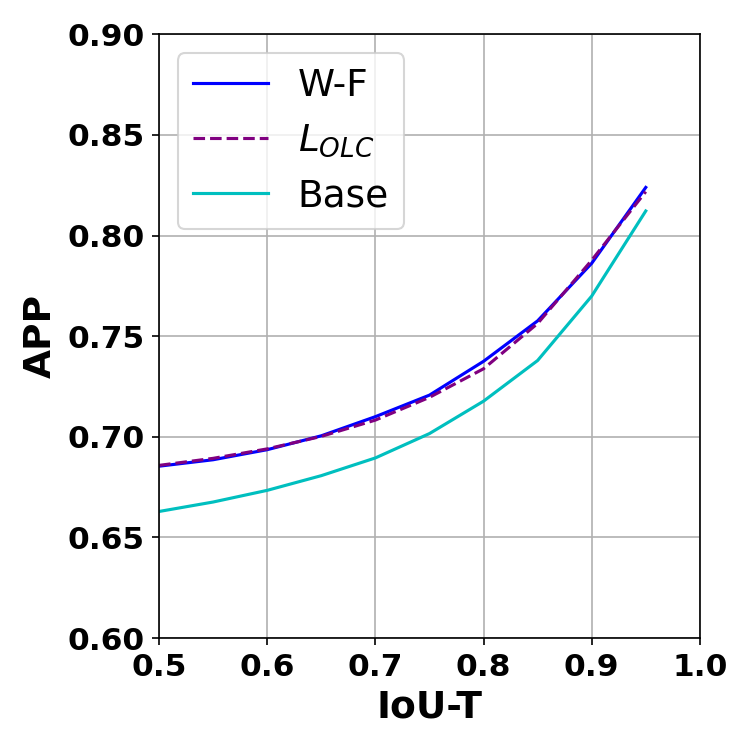}
    }
    \subfloat[]{
        \includegraphics[width=0.22\textwidth]{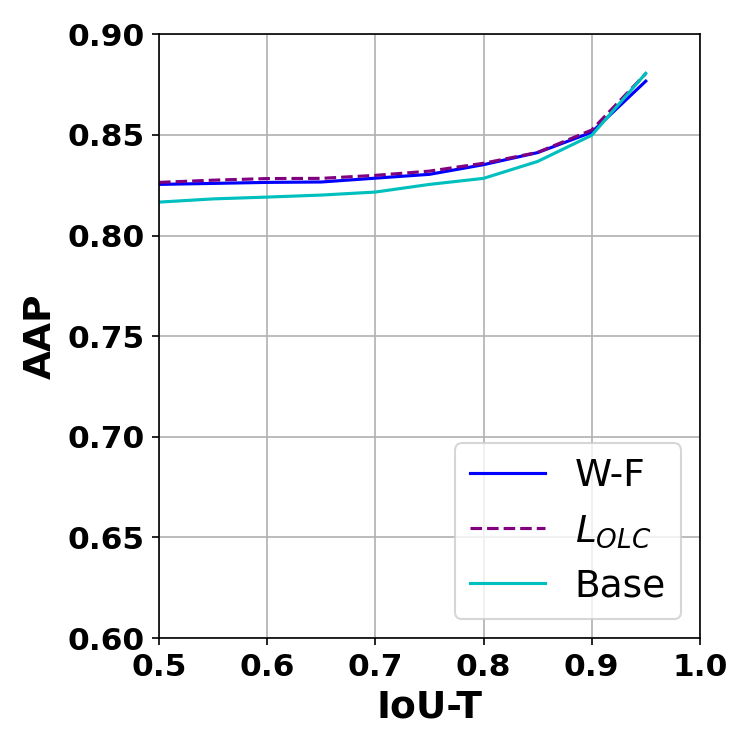}
    }
    \subfloat[]{
        \includegraphics[width=0.22\textwidth]{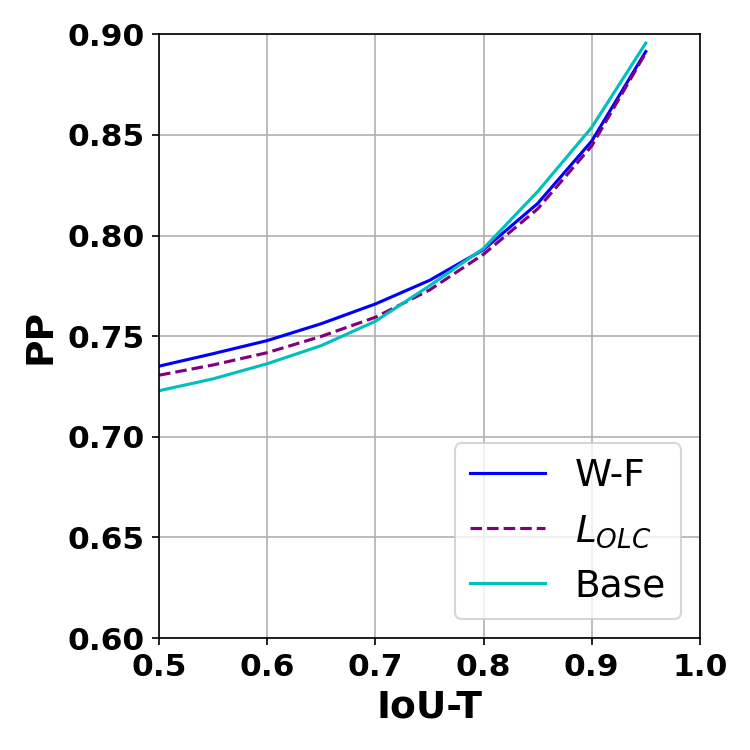}
    }       
    \caption{The effect of adding window-following(W-F) and $L_{OLC}$ on the detection performance in different IoU threshold. (a) ARP. (b) APP. (c) AAP. (d) PP.}
    \label{fig:ablation-PrIoU}
\end{figure*}
To verify the superiority of the DW mode images in wide-range vehicle detection, we compared our method of inputting DW images with the baseline method of inputting GW images and compared them on the P3DVD task. To make a fair comparison, the image resolution of the baseline input was adjusted to 1280x704 (i.e., the original image was scaled down 3 times), referred to as $\mbox{GW}'$, which is comparable in data size to DW.
\par
The comparative experimental results are presented in Table \ref{tab:comparative experiment on input image pattern}, where the vehicles are divided into four categories based on their pixel area size: small, medium, large, and all, following the evaluation criteria proposed by \cite{lin2014microsoft}, and the corresponding PR curves are shown in Fig. \ref{fig:P-R}. 
\par
The PR curves in Fig. \ref{fig:P-R} demonstrate the significant improvement of our method in detecting small and medium targets, while still achieving performance similar to the baseline in large category.
To be specific, as shown in Table \ref{tab:comparative experiment on input image pattern}, our method achieves 30.86$\%$ and 38.66$\%$ higher AP and AR than baseline in detecting small objects and achieves 20.04$\%$ and 20.05$\%$ higher in the medium category. The improvement benefits from CW image providing original distant vehicle information (typically falling in the small or medium category), leading to significantly better detection performance in the small and medium categories compared to the Baseline. Overall, our proposed method can significantly improve the detection performance of distant vehicles. It beats the baseline 13.51$\%$ AP and 14.40$\%$ AR in all category, while it only slightly increases the model parameters without adding computational cost.
\par
\begin{figure*}[htbp]
    \flushright
    \centering
    \includegraphics[width=0.85\textwidth]{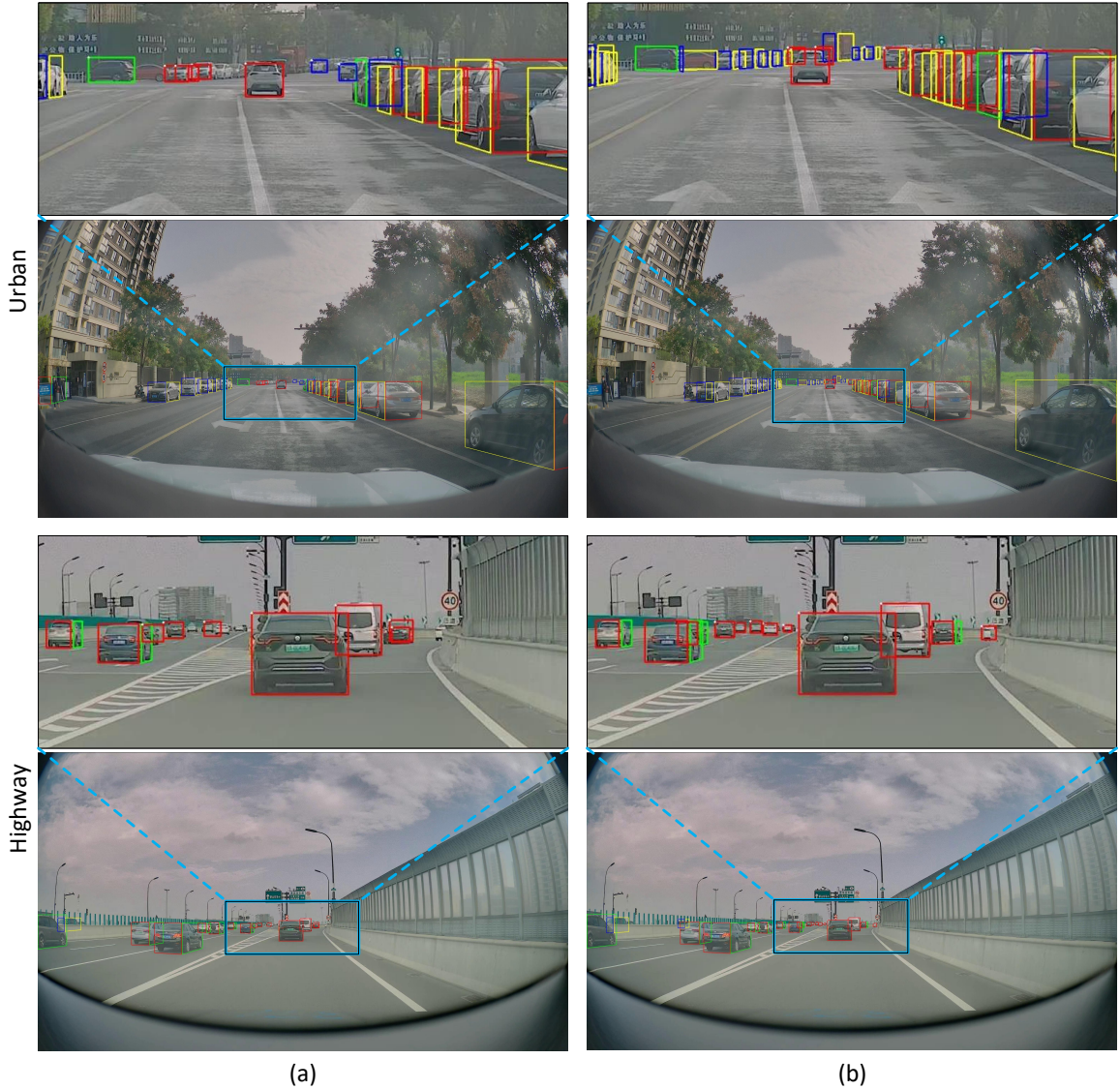}
    \caption{Distant vehicle detect comparison. The upper and lower rows in Urban and Highway scenes respectively visualize the detection results of CW and GW. (a) Baseline results. (b) Our results. The black rectangular box in the lower images represents the CW component of the DW images.}
    \label{fig:comparison}
\end{figure*}

To highlight the excellent performance of our method in detecting distant vehicles, we visually compared the prediction results of the Baseline and our method. As shown in Fig. \ref{fig:comparison}, in scenarios with complex traffic on city roads and open views on highways, our method can detect vehicles farther away compared to the Baseline, while also having comparable detection capabilities for larger nearby vehicles. Therefore, our method has the ability to detect vehicles over a wide range.


\subsection{Ablation Experiment}
During the model training stage, we designed the window-following image augmentation and the shape matching loss $L_{OLC}$ separately based on the characteristics of DW images and P3DVR to better train the model. To verify the effectiveness of these learning methods on the model, we conducted relevant ablation experiments.

\subsubsection{Contribution of Window-Following Image Augmentation}
As shown in Fig.\ref{fig:ablation-PrIoU}, it can be observed that window-following image augmentation significantly improves ARP and APP metrics, as well as AAP and PP metrics.
\par
The first and second rows of Table\ref{tab:Ablation Experiment} clearly show that window-following can comprehensively improve the detection performance of the model across all metrics, especially 0.83$\%$, 1.91$\%$, 3.88$\%$ and 3.49$\%$ higher in ARP, APP, AP, and AR. This is because window-following helps CW images always be extracted from the concentrated area of distant vehicles in the image, providing the model with more positive instances of small objects and thus more supervisory information during training. This is also the main reason why AP and AR can be greatly improved. However, the reasons for the relatively small improvements in PP, AAP, and ARP are that the classification of 8 vehicle poses is a simple task, and the model can learn it well with fewer training samples. Additionally, most of the distant vehicles in CW are in pose 0, as defined in Fig. \ref{fig:pose_definition}, which is recognized as an angle of -90° and a ratio of 1, so the method does not introduce much variation in training angles for the model.

\begin{figure*}[htbp]
    \centering
    \subfloat[]{
        \label{fig:ablation-kld-a}
        \includegraphics[width=0.45\textwidth]{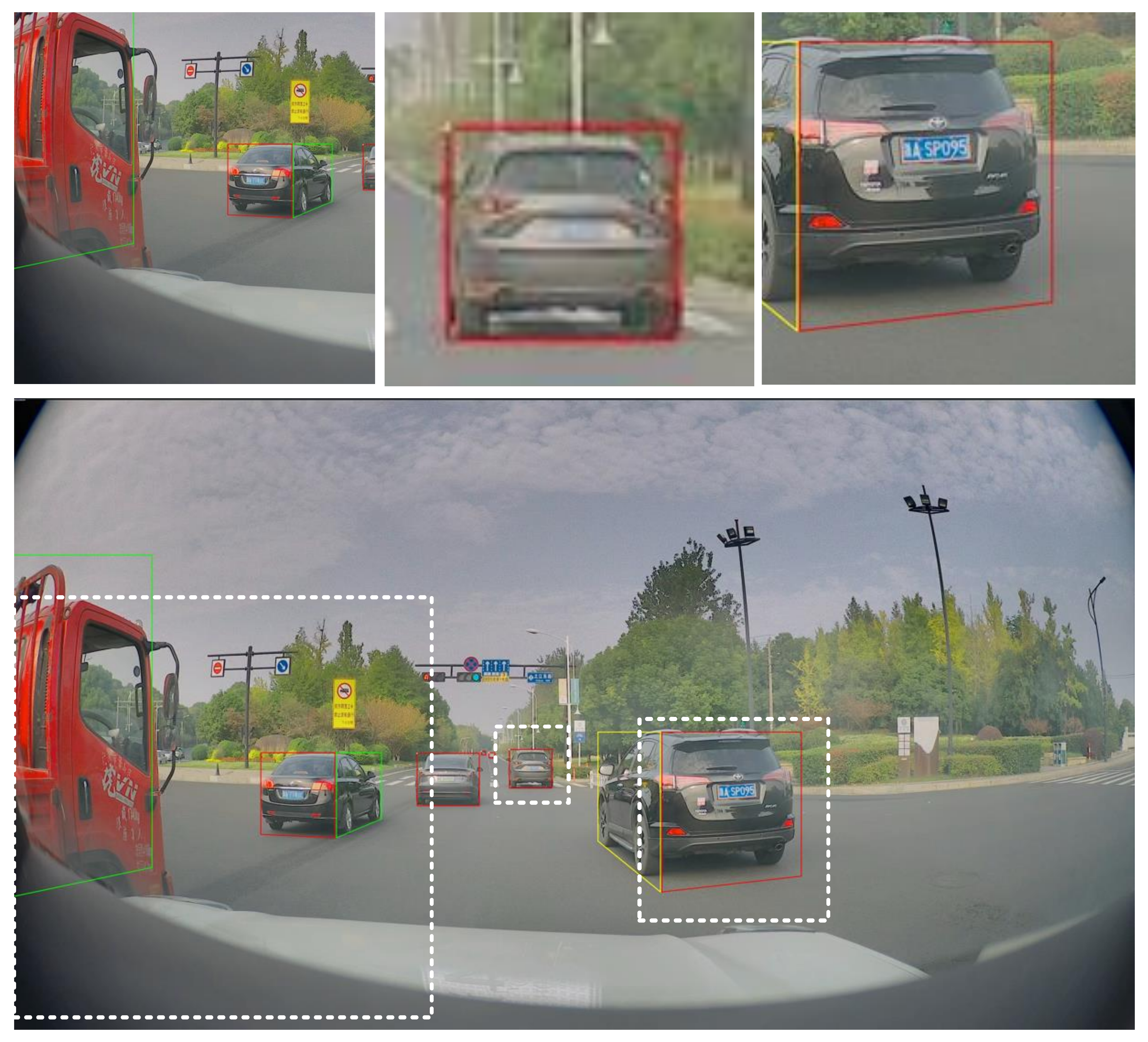}
        }
    \subfloat[]{
        \label{fig:ablation-kld-b}
        \includegraphics[width=0.45\textwidth]{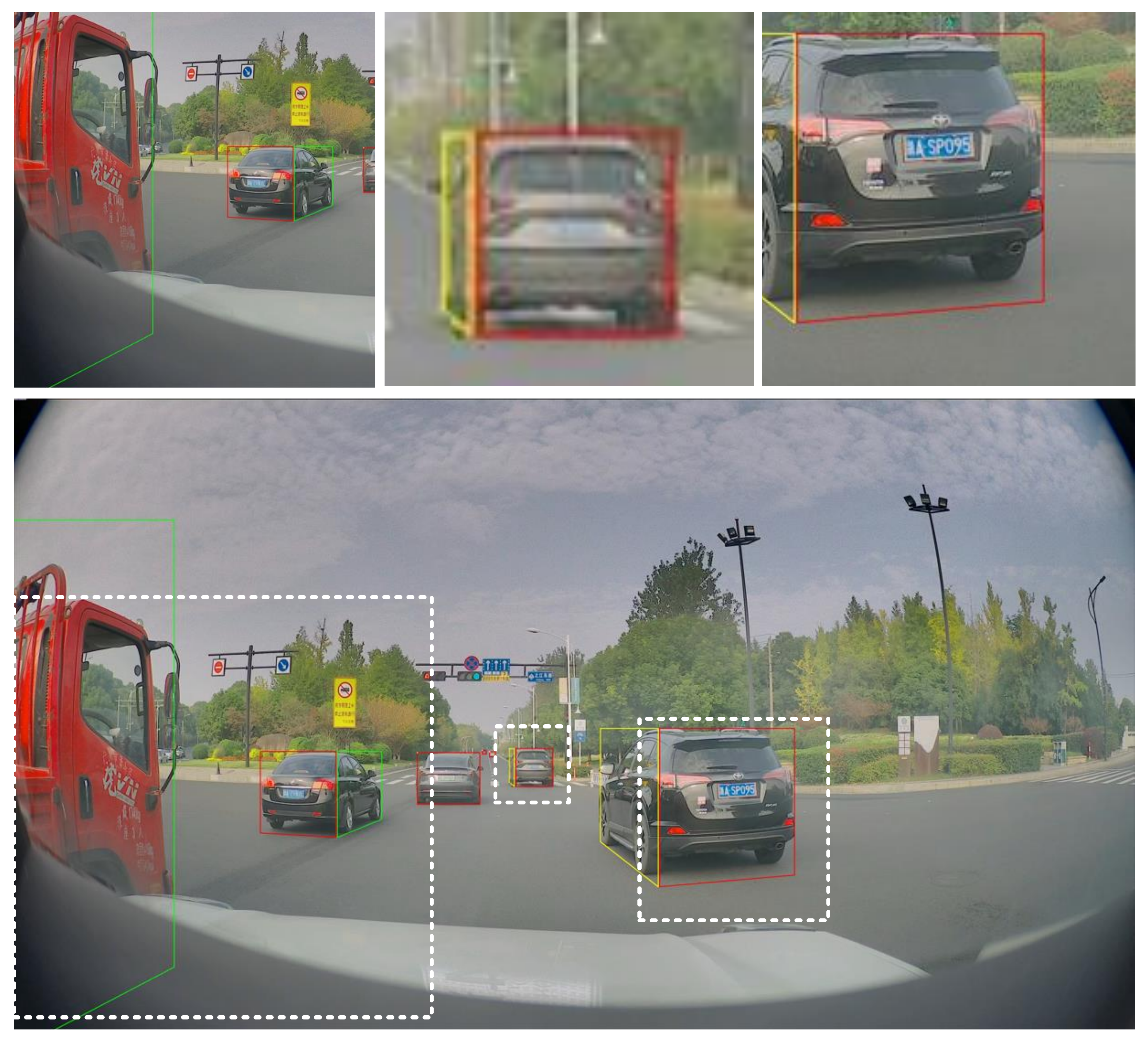}
    }
    \caption{Detection comparison between without and with $L_{OLC}$ supervision. (a) Training without $L_{OLC}$ supervision information. (b) Training with $L_{OLC}$ supervision information which obtains more accurate detection bboxes.}
    \label{fig:ablation-kld}
\end{figure*}
\subsubsection{Contribution of $L_{OLC}$}
Since there is no coupling relationship between data augmentation and loss functions, we directly added $L_{OLC}$ to the window-following image augmentation during model training. 
However, the addition of $L_{OLC}$ does not bring a very obvious improvement in evaluation metrics as shown in Fig.\ref{fig:ablation-PrIoU}.
To be specific, the comparison between the second and third rows of Table \ref{tab:Ablation Experiment} shows that adding $L_{OLC}$ during training, except for AAP, slightly weakened the model's detection performance on other metrics, such as ARP(-0.16$\%$), APP(-0.07$\%$) particularly PP(-0.42$\%$), compared with the results of our previous window-following method. From Equation \ref{Formula for calculating mean and variance of two-dimensional Gaussian distribution} and Fig. \ref{fig:Geometric Relationship between PSV3D targets and two-dimensional Gaussian distributions}, it can be seen that the participation of $pose$ in the calculation of $L_{OLC}$ is not high, and therefore, during training, the model is unable to supervise the predicted results of $pose$. Additionally, since the weights of the loss terms were not adjusted in the experimental process while controlling the variables, the model's attention to the learning of $pose$ attributes decreased.
\par 
To demonstrate the effectiveness of $L_{OLC}$ in improving model learning, we conducted visual comparisons. As shown in Fig. \ref{fig:ablation-kld-a}, the model trained without $L_{OLC}$ supervision produces P3DVR that does not fit the target instance well. In contrast, adding $L_{OLC}$ supervision helps the model to correct the P3DVR boundaries and better fit the vehicle contour. As shown in the top middle columns of Fig. \ref{fig:ablation-kld-b}, when the side view of the vehicle is less visible (the vehicle's yaw angle is close to 90°), SPL is highly sensitive to the angle, and slight disturbances can cause the P3DVR to lack realism. Adding $L_{OLC}$ supervision can help the model improve this phenomenon by predicting better $pose$ or higher-precision angles and midpoints of connection line between wheel ground points.

\subsection{Limitations}
\begin{figure}
    \centering
    \subfloat[]{
    \label{fig:Multiple vehicles overlap}
    \includegraphics[width=0.98\linewidth]{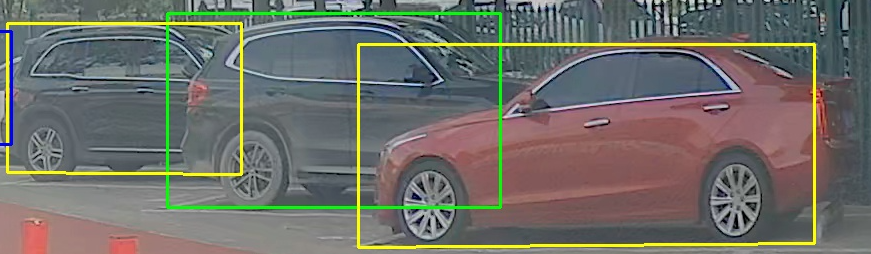}
    }\\
    \subfloat[]{
    \label{fig:too large}
    \includegraphics[width=0.43\linewidth]{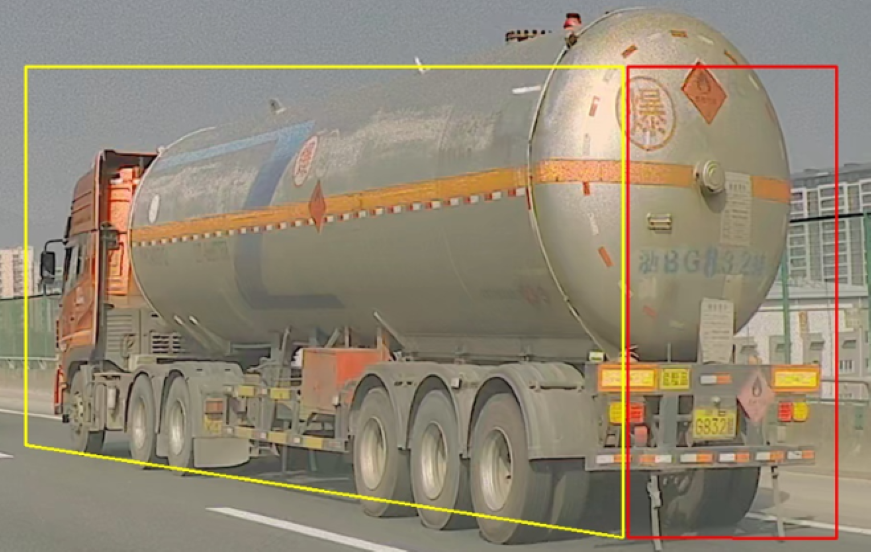}
    }
    \subfloat[]{
    \label{fig:wheels invisible}
    
    \includegraphics[width=0.52\linewidth]{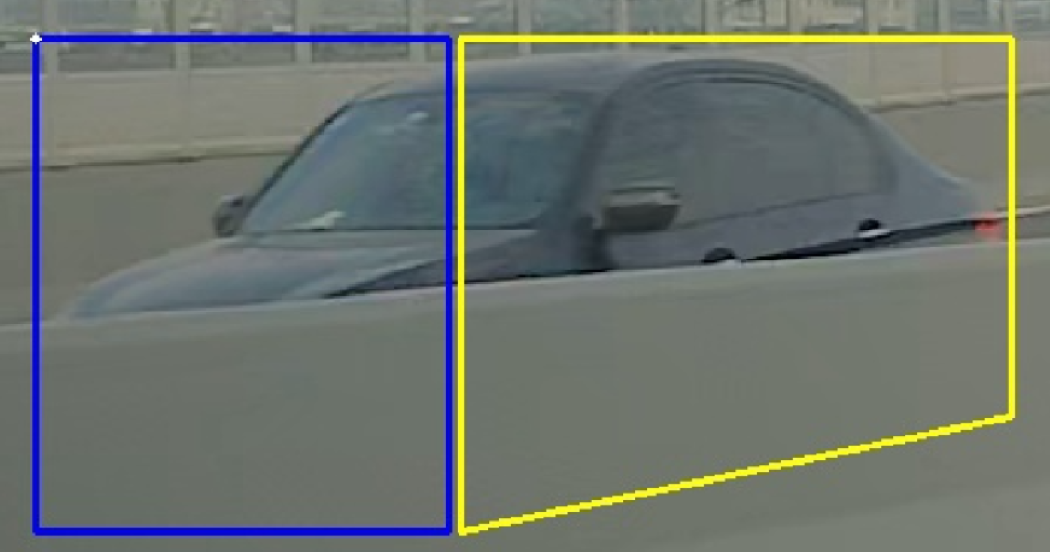}
    }
    \caption{Examples of method limitations. (a) Multiple vehicles overlap. (b) Too large vehicles. (c) Invisible wheels}
    \label{fig:limitations}
\end{figure}

Our method also has some limitations. When multiple vehicles overlap or the vehicle volume is too large, the P3DVR output by the model cannot accurately describe the vehicle contour, as shown in Fig.\ref{fig:Multiple vehicles overlap} and \ref{fig:too large}. When the front and rear wheels of the vehicle are occluded, the model cannot accurately predict the midpoint coordinates of the front and rear tire ground contact points so that the position of the SPL cannot be determined, as shown in Fig. \ref{fig:wheels invisible}.

\section{Conclusion}
We propose an efficient wide-range P3DVD method based on DW images and an efficient learning approach for the P3DVD model. DW images can provide a wealth of vehicle perception information for the model with limited resolution. To complete the P3DVD task, we designed a P3DVR specifically for vehicles to describe more geometric attributes of the vehicles. Based on the above two points, we accordingly designed window-following image augmentation and shape matching loss with joint constraints on object and SPL, helping the model to better utilize DW image information and learn more stable P3DVR predictions. Finally, we conducted corresponding experiments to demonstrate the effectiveness of our proposed method for wide-range P3DVD.
\par
Since the method proposed in this paper only presents a pseudo-3D representation of vehicles on the image and cannot directly obtain the actual geometric properties of the vehicles in 3D space, we will explore how the model can further predict the actual distance of the vehicle targets relative to the camera based on the P3DVD results in future research.

\bibliographystyle{IEEEtran}
\bibliography{ref}    

\begin{IEEEbiography}[{\includegraphics[width=1in,height=1.25in,clip,keepaspectratio]{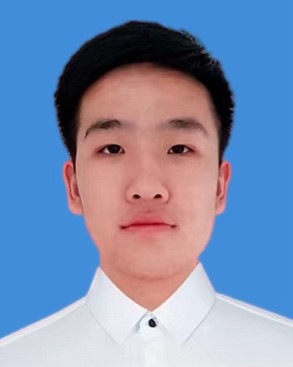}}]{Zhupeng Ye}
received the B.S. degree in automation from Ocean University of China, Qingdao, China, in 2021. He is currently pursuing the M.S. degree in artificial intelligence under the supervision of Dr. Zejian Yuan at Xi’an Jiaotong University, Xi’an, China. His research interests include computer vision and deep learning.
\end{IEEEbiography}

\begin{IEEEbiography}[{\includegraphics[width=1in,height=1.25in,clip,keepaspectratio]{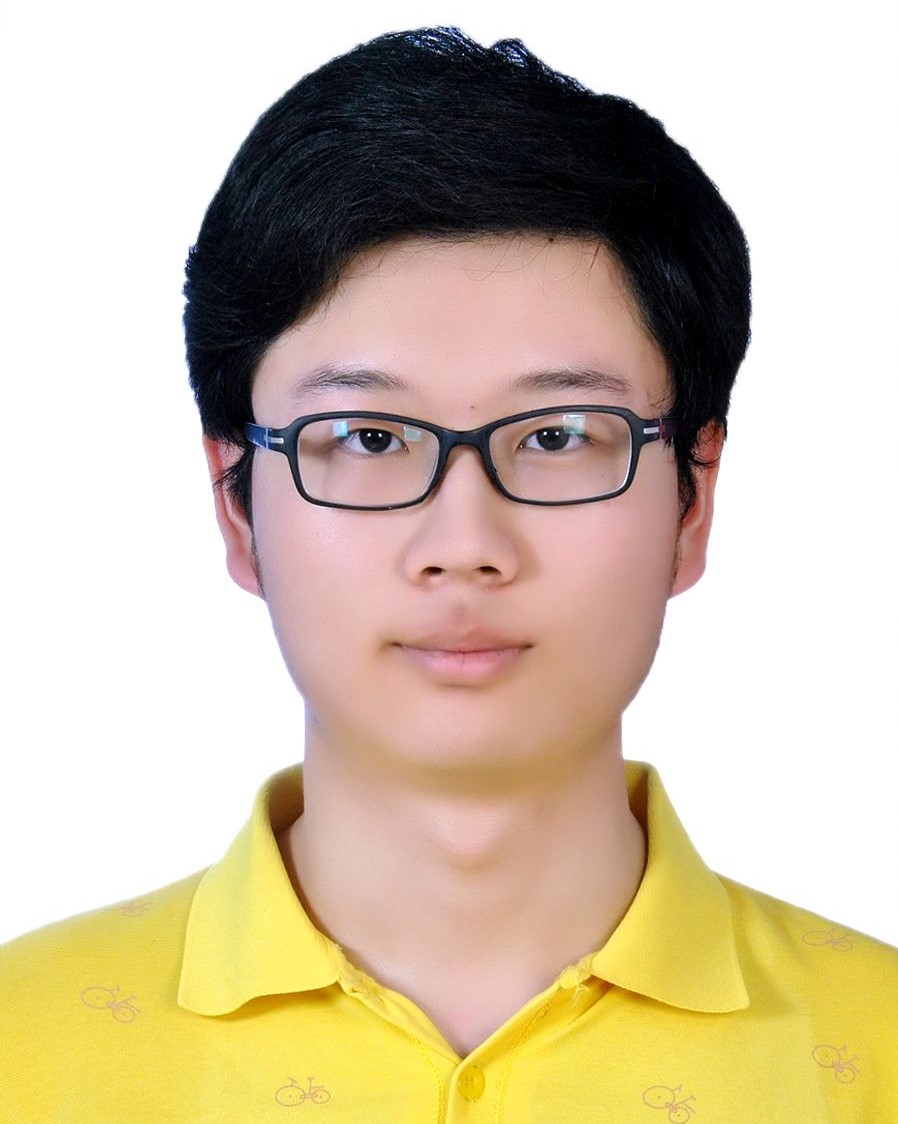}}]{Yinqi Li}
received the B.S. degree in artificial intelligence from Ocean University of China, Qingdao, China, in 2022. He is currently pursuing the M.S. degree in artificial intelligence under the supervision of Dr. Zejian Yuan at Xi’an Jiaotong University, Xi’an, China. His research interests include computer vision and video comprehension.
\end{IEEEbiography}

\begin{IEEEbiography}[{\includegraphics[width=1in,height=1.25in,clip,keepaspectratio]{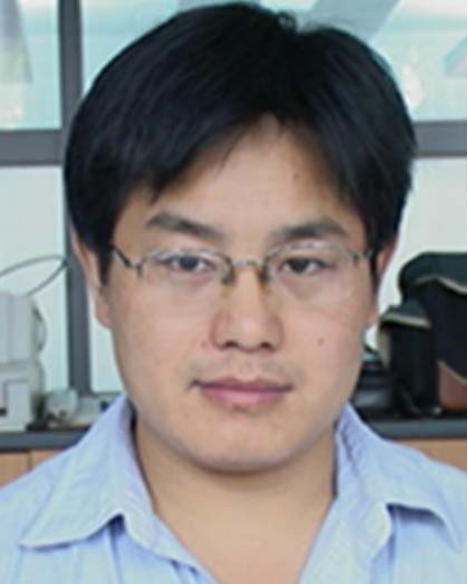}}]{Zejian Yuan}
(Member, IEEE) received the M.S. degree in electronic engineering from the Xi’an University of Technology, Xi’an, China, in 1999, and the Ph.D. degree in pattern recognition and intelligent systems from Xi’an Jiaotong University, China, in 2003. He is currently a Professor with the College of Artificial Intelligence, Xi’an Jiaotong University, and a member of the Chinese Association of Robotics. His research interests include image processing, pattern recognition, and machine learning in computer vision.

\end{IEEEbiography}

\vfill
\end{document}